\documentclass{article}

\usepackage{microtype}
\usepackage{graphicx}
\usepackage{bm}
\usepackage{booktabs}
\usepackage{colortbl}
\usepackage{multirow}
\usepackage{diagbox}

\usepackage{hyperref}
\usepackage{doi}

\usepackage[indLines=true,noEnd=true,commentColor=black]{algpseudocodex}
\usepackage[accepted]{icml2023}

\usepackage[subrefformat=parens]{subcaption}
\DeclareCaptionLabelSeparator{dot}{.~}
\usepackage{amsmath}
\usepackage{amssymb}
\usepackage{mathtools}
\usepackage{amsthm}
\usepackage{cancel}

\usepackage[capitalize]{cleveref}
\crefname{line}{line}{lines}

\makeatletter
\DeclareRobustCommand{\labelcrefrange}[2]{\@crefrangenostar{labelcref}{#1}{#2}}
\makeatother

\newcommand{\AutoGP}{\textsl{AutoGP}}

\theoremstyle{plain}
\newtheorem{theorem}{Theorem}
\newtheorem{proposition}[theorem]{Proposition}

\theoremstyle{definition}

\theoremstyle{remark}

\newtheorem*{remark*}{Remark}

\usepackage[textsize=tiny]{todonotes}

\usepackage{adjustbox}
\usepackage{comment}
\usepackage{graphicx}
\usepackage{layout}
\usepackage{microtype}
\usepackage{mdframed}
\usepackage{lipsum}
\usepackage[inline]{enumitem}

\usepackage{tikz}
\usepackage{tikz-qtree}
\usetikzlibrary{calc}
\usetikzlibrary{decorations.pathreplacing}
\usetikzlibrary{patterns}
\usetikzlibrary{shapes}

\DeclarePairedDelimiter\abs{\lvert}{\rvert}
\DeclarePairedDelimiter\set{\lbrace}{\rbrace}

\newcommand{\defas}{\coloneq}
\newcommand{\diff}{\mathop{}\!\mathrm{d}}
\newcommand{\simiid}{\overset{\rm iid}{\sim}}
\newcommand{\Dist}[2][]{\mbox{#1{#2}}}
\newcommand{\probe}{\psi}

\algnewcommand{\LineComment}[1]{\State {\footnotesize \(\triangleright\) #1}}

\begin{document}

\twocolumn[
\icmltitle{Sequential Monte Carlo Learning for Time Series Structure Discovery}



\icmlsetsymbol{equal}{*}

\begin{icmlauthorlist}
\icmlauthor{Feras A.~Saad}{cmu,google}
\icmlauthor{Brian J.~Patton}{google}
\icmlauthor{Matthew D.~Hoffmann}{google}
\icmlauthor{Rif A.~Saurous}{google}
\icmlauthor{Vikash K.~Mansinghka}{mit,google}
\end{icmlauthorlist}

\icmlaffiliation{cmu}{Carnegie Mellon University}
\icmlaffiliation{google}{Google Research}
\icmlaffiliation{mit}{Massachusetts Institute of Technology}
\icmlcorrespondingauthor{Feras Saad}{fsaad@cmu.edu}

\icmlkeywords{time series, structure learning, sequential Monte Carlo}

\vskip 0.3in
]

\printAffiliationsAndNotice{}  

\begin{abstract}
This paper presents a new approach to automatically discovering
accurate models of complex time series data.
Working within a Bayesian nonparametric prior over a symbolic space of
Gaussian process time series models, we present a novel structure
learning algorithm that integrates sequential Monte Carlo (SMC) and
involutive MCMC for highly effective posterior inference.
Our method can be used both in ``online'' settings, where new data is
incorporated sequentially in time, and in ``offline'' settings, by
using nested subsets of historical data to anneal the posterior.
Empirical measurements on real-world time series show that our method
can deliver 10x--100x runtime speedups over previous MCMC and
greedy-search structure learning algorithms targeting the same model family.
We use our method to perform the first large-scale evaluation of
Gaussian process time series structure learning on a prominent
benchmark of 1,428 econometric datasets.
The results show that our method discovers sensible models that
deliver more accurate point forecasts and interval forecasts over
multiple horizons as compared to widely used statistical and neural
baselines that struggle on this challenging data.
\end{abstract}

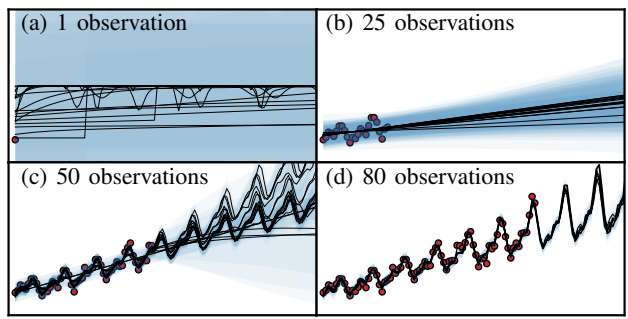
\begin{figure}[t]

\begin{tikzpicture}
\captionsetup{font={normalfont},labelfont={bf}}
\node[name=pic]{\includegraphics[width=\linewidth]{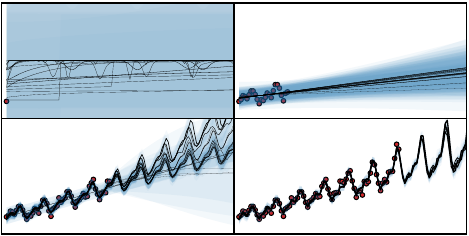}};
\node[name=a,
  text width=27.5mm,
  at={(pic.north west)},
  anchor=north west,
  xshift=2mm,
  yshift=-2mm,
  draw=none,
  fill=none,
  inner sep=0pt,
  outer sep=0pt,
]{
\subcaptionlistentry{}
\label{fig:real-time-1}
\subcaptiontext*{1 observation}};

\node[name=b,
  text width=27.5mm,
  right={13mm of a},
  draw=none,
  fill=none,
  inner sep=0pt,
  outer sep=0pt,
]{
\subcaptionlistentry{}
\label{fig:real-time-25}
\subcaptiontext*{25 observations}};

\node[name=c,
  text width=27.5mm,
  below={20.5mm of a.west},
  anchor=west,
  draw=none,
  fill=none,
  inner sep=0pt,
  outer sep=0pt,
]{
\subcaptionlistentry{}
\label{fig:real-time-50}
\subcaptiontext*{50 observations}
};

\node[name=c,
  text width=27.5mm,
  draw=none,
  fill=none,
  inner sep=0pt,
  outer sep=0pt,
  at={(c.center -| b.center)},
  anchor=center
]{
\subcaptionlistentry{}
\label{fig:real-time-80}
\subcaptiontext*{80 observations}
};
\end{tikzpicture}
\caption{Real-time discovery of time series structure.
Red dots show observed data so far, solid lines show mean forecasts
from a weighted ensemble of learned symbolic model structures, where
the thickness of a line is proportional to its corresponding model weight; blue
regions show 95\% prediction intervals.
\subref{fig:real-time-1} Broad uncertainty.
\subref{fig:real-time-25} Dominant linear trend.
\subref{fig:real-time-50} Multimodal posterior,
including linear trend with additive or with multiplicative seasonality.
\subref{fig:real-time-80} Linear trend with multiplicative seasonality.
}
\label{fig:real-time}
\vspace{-15pt}
\end{figure}

\section{Introduction}
\label{sec:intro}

Many applications depend on models that can explain and predict
time series data.
A key challenge in time series modeling is the substantial expertise
needed to design statistical models that capture complex temporal
patterns such as changepoints, heteroskedastic noise, additive and
multiplicative seasonal effects, and higher-order
autocorrelations~\citep[Chapters 12--13]{hyndman2021}.
Even experts who know how to construct sophisticated time series
models may have trouble deciding what types of patterns to include
when working with a given dataset.
Motivated by these challenges, this work addresses the problem of
automatically discovering models of time series data
that exhibit a wide range of patterns which are a-priori unknown.

Given observations $\set{(t_i, y_i)}_{i=1}^n$ of time points $t_i$ and
time series values $y_i$, we focus on learning an unknown real
function $f$ and i.i.d.~noise process $\epsilon$ such that
$y(t) = f(t) + \epsilon(t), t \in \mathbb{R}$.
These so-called ``pure time series models'', which are univariate in nature
and do not include exogenous variables, are used in many disciplines.
In econometrics, for example, the function $f$ is intended to capture
essential empirical features of the observed data that are believed to
arise from an unknown macroeconomic structural model whose exogenous
variables are too complex or impossible to accurately
specify~\citep[Chapter 5]{brooks2008}.

\Cref{fig:real-time} shows an example data stream and the evolution of
various hypotheses about the unknown function $f$ that our method
explores as it encounters new observations.
The plots illustrate some central features of our
approach, which include:
\begin{enumerate*}[label=(\roman*)]
\item adapting both the structure and parameters of $f$ according to
patterns in the data observed so far; and
\item maintaining not one hypothesis about $f$ but a weighted
collection of hypotheses that together capture uncertainty over
the unknown time series structure and parameters.
\end{enumerate*}

\paragraph{Overview}
To automatically discover time series models, we stochastically sample
symbolic expressions in the Gaussian process family for nonparametric
regression introduced in \citet{duvenaud2013}, which we specialize to
univariate time series data~\citep{roberts2013}.
Whereas \citet{duvenaud2013} use a greedy-search algorithm for
structure learning, our approach is based on fully Bayesian inference
in a probabilistic generative model over symbolic descriptions of
temporal patterns (i.e., Gaussian process covariance functions),
numeric parameters, and observed data.
This representation enables us to develop a novel sequential Monte
Carlo (SMC) algorithm for posterior inference that naturally handles
streaming data and coherently characterizes joint uncertainty about
latent model structure and parameters.

\paragraph{Key Results} The SMC algorithm introduced in this paper can
deliver 10x--100x improvements in runtime vs.~forecasting accuracy
profiles for many real-world datasets (\cref{fig:runtime}) over the
greedy-search method in \citet{duvenaud2013} (which finds a single
``best-fit'' model structure and parameters) and over the MCMC sampling
method in~\citet{saad2019}.
We further evaluate our method on 1,428 monthly datasets from
M3~\citep{makridakis2000} against 9 statistical and neural
baselines~\citep{hyndman2008,hyndman2008ets,fiorucci2016,golyandina2018,taylor2018,lindemann2021,lim2021},
many of which struggle to capture the breadth of patterns in the
data~\citep{makridakis2018}.
The results show that our structure-learned models produce more
accurate point forecasts (improvement $\ge 14\%$) and interval
forecasts (improvement $\ge 6\%$) across 1--18 step horizons
(\cref{fig:m3-eval,fig:m3-explore}), with a runtime per dataset that
is comparable to the most competitive baselines
(\cref{tbl:m3-runtime}).

\paragraph{Outline}
The rest of the paper is structured as follows:
\cref{sec:background} reviews time series modeling using Gaussian processes;
\cref{sec:smc} presents our SMC structure learning method;
\cref{sec:evaluation} contains the evaluation;
\cref{sec:related} discusses related work; and
\cref{sec:conclusion} offers closing remarks.

\section{Gaussian Process Time Series Models}
\label{sec:background}

\paragraph{Preliminaries}
Let $D \defas (\mathbf{t}, \mathbf{y})$ be a dataset with
$n$ time points $\mathbf{t} \defas (t_1,\dots,t_n)$
and observations $\mathbf{y} \defas (y_1,\dots,y_n)$.
As the true function that generated the data is unknown,
we use a Gaussian process function prior
$f \sim \Dist{GP}(0, k_\theta)$ and assume an additive observation noise model, so
that $y_i = f(t_i) + \epsilon_i$, where $k_\theta \defas (k,\theta)$ denotes a
covariance kernel $k$ with parameters
$\theta = (\theta_1, \dots, \theta_{d(k)})\,{\in}\,\Theta_k\,{\subset}\,\mathbb{R}^{d(k)}$;
$d(k)$ is the number of continuous parameters in $k$;
and $\epsilon_i\,{\simiid}\,N(0, \eta)$.
The probabilistic generative model
$P_{\mathbf{t}}(y; k_\theta)$ is then
\begin{align}
\label{eq:gp-model-1} [f(t_1), \dots, f(t_n)] &\sim \Dist{MultivariateNormal}(0, k_\theta(\mathbf{t}))\\
\label{eq:gp-model-2} y_i &\sim \Dist{Normal}(f(t_i), \eta), \quad i \in [n].
\end{align}

\begin{figure}[t]
\includegraphics[width=\linewidth]{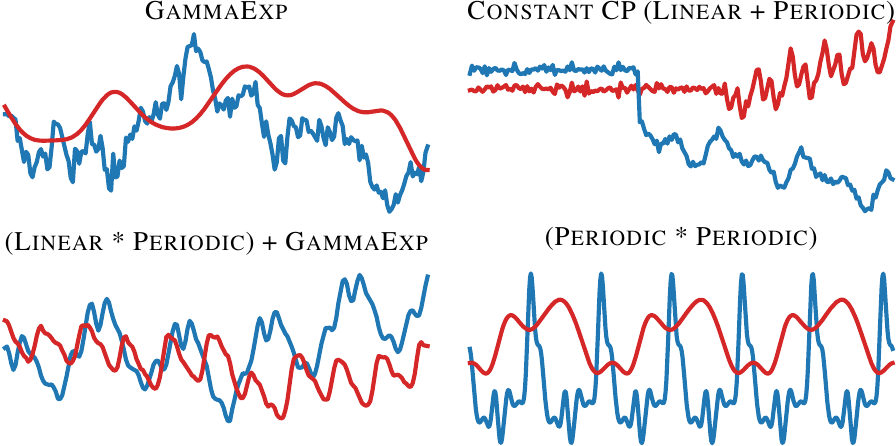}
\caption{Example time series
patterns that can be expressed in the Gaussian process model family
(\crefrange{eq:pcfg-1}{eq:GP-prior-5}), including random-walk
evolution, smooth variation, trends with additive and multiplicative
seasonality, changepoints (\textsc{CP}), and multiple seasonality. For each kernel
structure $k$, colored lines show draws of random functions
$f \sim \mathrm{GP}(0,k_\theta)$ using different parameters
$\theta \in \mathbb{R}^{d(k)}$.}
\label{fig:gp-tutorial}
\end{figure}

\paragraph{A Flexible Modeling Language}
While the model~\labelcrefrange{eq:gp-model-1}{eq:gp-model-2} appears
simple on the surface, Gaussian processes define a flexible
class of distributions.
The kernel structure $k$ and parameters $\theta$ of the
covariance function $k_\theta$ together dictate the properties of the random
function $f$.
\Citet{duvenaud2013} introduced a language $\mathcal{L}$
for expressing a wide range of Gaussian process patterns, which we
restrict to time series data.
Kernel structures $k \in \mathcal{L}$ are specified compositionally from
simpler parts, using a context-free grammar (CFG):
\begin{align}
\label{eq:pcfg-1}
B      &\Coloneq \textsc{Linear} \mid \textsc{Periodic} \mid \textsc{GammaExp} \mid \dots \\
\label{eq:pcfg-2}
\oplus &\Coloneq \textsc{+} \mid \textsc{*} \mid \textsc{CP} \\
\label{eq:pcfg-3}
k      &\Coloneq B \mid \textsf{(}k_1 \oplus k_2\textsf{)}.
\end{align}
The symbol $B$ denotes a finite set of ``base'' kernels that
can be composed through binary operators $\oplus$, namely
addition ($\textsc{+}$), pointwise multiplication ($\textsc{*}$), and
changepoints ($\textsc{CP}$)~\citep[Chapter 4]{rasmussen2006}.
Base kernels $b \in B$
and the changepoint operator $\textsc{CP}$
are associated with parameter spaces $\Theta_b \subset \mathbb{R}^{d(b)}$,
$\Theta_{\textsc{CP}} \subset \mathbb{R}^2$, respectively.
\Cref{fig:gp-tutorial} shows draws $f \sim \mathrm{GP}(0, k_\theta)$,
for various $k_\theta$.

\paragraph{A Prior Over Language Expressions}
To automate the process of discovering covariance functions
$k_\theta$ for a given dataset, we treat all unknown quantities as latent
variables in a probabilistic generative model
$P_{\mathbf{t}}(k, \theta, \eta, \mathbf{y})$ defined by
\begin{align}
\label{eq:GP-prior-1}
k &\sim \Dist{PCFG},  \mbox{c.f.~\labelcref{eq:pcfg-1,eq:pcfg-2,eq:pcfg-3}} \\
\label{eq:GP-prior-2}
[\theta_1, \dots, \theta_{d(k)}] &\simiid \Dist{LogNormal}(0, 1) \\
\label{eq:GP-prior-3}
\eta &\sim \Dist{InverseGamma}(1, 1) \\
\label{eq:GP-prior-4}
[f(t_1), \dots, f(t_n)] &\sim \Dist{MultivariateNormal}(0, k_\theta(\mathbf{t}))  \\
\label{eq:GP-prior-5}
y_i &\sim \Dist{Normal}(f(t_i), \eta), \quad i \in [n].
\end{align}
In \cref{eq:GP-prior-1}, ``PCFG'' denotes a prior over $\mathcal{L}$
obtained by assigning a probability to each production rule in the
CFG~\labelcrefrange{eq:pcfg-1}{eq:pcfg-3}.
To ease notation, we will use the symbol $\varphi \defas (\theta, \eta)$
to denote the numeric kernel parameters and noise variance.

\paragraph{Problem Formulation}
In \crefrange{eq:GP-prior-1}{eq:GP-prior-5}, the posterior
\begin{align}
&P_{\mathbf{t}}(k,\varphi \mid \mathbf{y})
  = \frac
    {P_{\mathbf{t}}(k,\varphi, \mathbf{y})}
    {\sum_{k \in \mathcal{L}} \int_{\Theta_k \times [0,\infty]}
    P_{\mathbf{t}}(k,\varphi, \mathbf{y})\diff{\varphi}
    },
\label{eq:target-posterior}
\end{align}
over $(k,\varphi)$ given $\mathbf{y}$
is defined on a transdimensional space,
\begin{align}
&\mathrm{support}\left( P_{\mathbf{t}}(k,\varphi \mid \mathbf{y}) \right)
\subset \bigcup_{k \in \mathcal{L}} \big[ \set{k} \times \Theta_k \times \mathbb{R} \big].
\label{eq:target-support}
\end{align}
Our goal is to generate a set
$\set{(w^i, (k^i, \varphi^i))}_{i=1}^M$
of $M \ge 1$ approximate posterior samples
$(k^i, \varphi^i)\sim P_{\mathbf{t}}(k,\varphi \mid \mathbf{y})$
and weights $w^i\,{>}\,0$ that can be used to solve queries,
i.e., to compute posterior expectations of test functions $\probe$,
\begin{align}
\mathbb{E}_{P_{\mathbf{t}}} \left[ \probe(k,\varphi,\mathbf{t},\mathbf{y}) \mid \mathbf{y} \right]
\approx \sum_{i=1}^M \frac{w^i}{\textstyle\sum_{j} w^j} \probe(k^i,\varphi^i,\mathbf{t},\mathbf{y}).
\label{eq:liquescence}
\end{align}
The query may be about the posterior probability
of the presence of a given temporal structure in the data, e.g.,
\begin{align}
P_{\mathbf{t}}(\mathsf{Periodic} \in k \mid \mathbf{y})
= \mathbb{E}_{P_{\mathbf{t}}}\left[\mathbf{1}[\mathsf{Periodic} \in k] \mid \mathbf{y} \right],
\end{align}
or a prediction for new data $\mathbf{y}_*$ at time points $\mathbf{t}_*$, e.g.,
\begin{align}
&\mathbb{E}_{P_{\mathbf{t}, \mathbf{t}_*}}\left[ f(\mathbf{t}_*) \mid \mathbf{y} \right]
  = \mathbb{E}_{P_{\mathbf{t}}}\left[\probe^{\rm mean}_{\mathbf{t}_*}(k,\varphi,\mathbf{t},\mathbf{y}) \mid \mathbf{y} \right],
  \label{eq:query-mean}
  \\
&\quad \probe^{\rm mean}_{\mathbf{t}_*} \defas
  \int_{-\infty}^{\infty}
    \mathbf{u_*}
    P_{\mathbf{t},\mathbf{t_*}}(f(\mathbf{t_*}) = \mathbf{u}_* \mid k, \varphi, \mathbf{y})
    \diff{\mathbf{u}_*};
    \notag
\\
&P_{\mathbf{t}, \mathbf{t}_*}(\mathbf{y_*} \le \mathbf{u} \mid \mathbf{y})
  = \mathbb{E}_{P_{\mathbf{t}}}\left[\probe^{\rm CDF}_{\mathbf{t}_*,\mathbf{u}}(k,\varphi,\mathbf{t},\mathbf{y}) \mid \mathbf{y} \right],
  \label{eq:query-cdf} \\
&\quad \probe^{\rm CDF}_{\mathbf{t}_*,\mathbf{u}} \defas
  \int_{-\infty}^{\mathbf{u}}
    P_{\mathbf{t},\mathbf{t_*}}(\mathbf{y}_* = \mathbf{u}_* \mid k, \varphi, \mathbf{y})
    \diff{\mathbf{u}_*}.
    \notag
\end{align}
Since $P_{\mathbf{t},\mathbf{t_*}}(f(\mathbf{t_*}) \mid k, \varphi, \mathbf{y})$
in \cref{eq:query-mean,eq:query-cdf} is also a multivariate normal distribution,
the test function $\probe^{\rm mean}_{\mathbf{t}_*}$ can
be analytically computed in closed form (as can other moments, see
\citet[Chapter 2.2]{rasmussen2006}), whereas
$\probe^{\rm CDF}_{\mathbf{t}_*, \mathbf{u}}$ can be computed numerically
with high precision~\citep{genz1992}.
Generally speaking, evaluating $\probe(k,\varphi,\mathbf{t},\mathbf{y})$ pointwise is
``easy''; whereas computing its posterior expectation as in
\crefrange{eq:query-mean}{eq:query-cdf} requires the
particle collection $\set{(w^i, (k^i, \varphi^i))}_{i=1}^M$
for the estimator~\labelcref{eq:liquescence}.

\section{A Sequential Monte Carlo Sampler for Time Series Structure Learning}
\label{sec:smc}

In this section we briefly review sequential Monte Carlo (SMC)
sampling and describe a novel SMC sampler for inferring a weighted
collection of kernel structures and parameters that approximate the
posterior distribution~\labelcref{eq:target-posterior}.

\paragraph{Background}

SMC methods~\citep{delmoral2007} are designed to produce approximate
samples from a sequence $\pi_0, \pi_1, \dots, \pi_T$ of
probability distributions, where each $\pi_j$ is defined on a measurable space
$(X_j, \mathcal{X}_j)$ can be evaluated pointwise up to a normalizing constant,
$\pi_j(x) = \gamma_j(x)/Z_j$ $(j=0,\dots,T; Z_0 \equiv 1)$.
The output of an SMC sampler at step $j$ (for $j= 0,\dots,T$)
is a weighted ``particle collection'' $\set{(w^i_{j},x^i_{j})}_{i=1}^M$, such that
each random pair $(w^i_j,x^i_j)$ is ``properly weighted'' \citep{liu1998}
for $\pi_j$, meaning
\begin{align}
\mathbb{E}\left[w^i_j\ \probe(x^i_j)\right] = Z_j \mathbb{E}_{\pi_j}\left[\probe(x)\right],
&& (\varphi \in X_j \to \mathbb{R}).
\label{eq:properly-weighted}
\end{align}
As \cref{eq:properly-weighted} implies that $\mathbb{E}[w^i_j] = Z_j$,
the particles can be used to obtain biased but consistent estimators
of expectations $\mathbb{E}_{\pi_j}\left[\probe(x)\right]$ via ratio estimation.
Assuming that $\pi_0$ is tractable to sample from and $Z_0 = 1$, the first step in SMC
is to generate $x^i_0 \sim \pi_0$ and $w^i_0 \gets 1$ ($i=1,\dots,M$),
which gives properly weighted pairs for $\pi_0$.
At step $j$, the particles $\set{(w^i_{j-1}, x^i_{j-1})}_{i=1}^M$ are
evolved through a two-step process
\begin{align}
\label{eq:smc-step-1}
x^i_j &\sim K_j(x^i_{j-1}, \cdot) \\
\label{eq:smc-step-2}
w^i_j &\gets w^i_{j-1} \left[
  \frac{\gamma_{j}(x^i_{j}) L_{j-1}(x^i_{j}, x^i_{j-1}) }{\gamma_{j-1}(x^i_{j-1}) K_j(x^i_{j-1}, x^i_j)}
\right],
\end{align}
where $K_j: X_{j-1} \times \mathcal{X}_j \to [0,1]$ and $L_{j-1}:
X_{j} \times \mathcal{X}_{j-1} \to [0,1]$ are Markov kernels
(whose densities we also refer to as $K_j$ and $L_{j-1}$).
These kernels are typically designed on a per-problem basis.
Subject to relatively mild conditions, the new particle collection
$\set{(w^i_j,x^i_j)}_{i=1}^M$ obtained by \crefrange{eq:smc-step-1}{eq:smc-step-2}
is properly weighted for $\pi_j$.
To help prevent particle collapse, if the weights
$w_j^{1:M}$ at step $j$ become skewed, each particle $x^i_j$ is
resampled to take value $x^\iota_j$ with (relative) probability $w^\iota_j$ $(\iota=1,\dots,M)$
and its weight is reset to the mean weight: $w^i_j \gets (w^1_j + \dots + w^M_j) / M$.

\subsection{Sequential Bayesian Structure Learning}

To leverage SMC for inferring the posterior
$P_\mathbf{t}(k,\varphi\,{\mid}\,\mathbf{y})$ in
\cref{eq:target-posterior} we use a sequential Bayesian reasoning
approach.
The observations $D = (\mathbf{t}, \mathbf{y})$
are partitioned into $T$ subsets $D_j \defas (\mathbf{t}_j, \mathbf{y}_j)$
with
$\mathbf{t}_j \defas t_{j,1:N_j}$, and
$\mathbf{y}_j \defas y_{j,1:N_j}$
($j = 1, \dots, T; N_j > 0$).
We let $\mathbf{t}_{1:j} \defas \cup_{s=1}^j \mathbf{t}_s$ and analogously
for $\mathbf{y}_{1:j}$, so
$(\mathbf{t}, \mathbf{y}) \equiv (\mathbf{t}_{1:T}, \mathbf{y}_{1:T})$.
In online learning problems (e.g.,~\cref{fig:real-time}) the $D_j$
correspond to streaming new data and in offline problems
to batches of previous data used to anneal the full
posterior~\labelcref{eq:target-posterior}.
The sequence of target distributions for SMC is then
$P_{\mathbf{t}_{1:j}}(k,\varphi\,{\mid}\,\mathbf{y}_{1:j})$,
$j=0,\dots,T$.

As all these target distributions are defined on the same
transdimensional space~\labelcref{eq:target-support}, we use a
\textit{reweight}-\textit{resample}-\textit{rejuvenate} SMC method
(\cref{alg:smc}) that is similar to the iterated batched importance
sampler (IBIS) of \citet{chopin2002}.
In the \textit{reweight} step (\crefrange{alg:smc-reweight-1}{alg:smc-reweight-2}), the target is updated from
$\pi_{j-1} \defas P_{\mathbf{t}_{1:j-1}}(k,\varphi\mid \mathbf{y}_{1:j-1})$
to
$\pi_{j} \defas P_{\mathbf{t}_{1:j}}(k,\varphi\mid \mathbf{y}_{1:j})$,.
For \cref{eq:smc-step-1}, the particle $(k^i_{j-1}, \varphi^i_{j-1})$ is evolved through a deterministic
``copy'' kernel $K_j$, i.e, $(k^i_j, \varphi^i_j) \sim \delta_{(k^i_{j-1}, \varphi^i_{j-1})}$,
so that the corresponding reverse kernel $L_{j-1}$ is an atom at $(k^i_{j}, \varphi^i_{j})$.
The incremental weight (bracketed term in~\cref{eq:smc-step-2}) is then
$P_{\mathbf{t}_{1:j}}(\mathbf{y}_j \mid k^i_j, \varphi^i_j, \mathbf{y}_{1:j-1})$.

\bgroup
\setlength{\textfloatsep}{0pt}
\begin{algorithm}[t]
\caption{SMC Structure Learning via Data Annealing}
\label{alg:smc}
\algrenewcommand\algorithmicindent{.75em}
\begin{algorithmic}[1]
\Require{
\begin{tabular}[t]{@{}l}
Dataset sequence $D_j \defas (\mathbf{t}_i, \mathbf{y}_j), j=1,\dots,T;$ \\
No.~of particles $M > 0$, rejuv steps $N_{\mathrm{rejuv}} \ge 0$. \\[-5pt]
\end{tabular}
}
\Ensure{
Properly weighted samples
  $\set{(w^i, (k^i, \varphi^i))}$
  for posterior distribution $P_{\mathbf{t}_{1:T}}(k, \varphi \mid \mathbf{y}_{1:T})$.
}
\LComment{\footnotesize Repeat operations indexed by $i$ over $i=1,\dots,M$}
\State $(k^i_0, \varphi^i_0) \sim P(k,\varphi)$; \quad $w^i_0 \gets 1$
\For{$j = 1, \dots, T$}
  \State $(k^i_j, \varphi^i_j) \gets (k^i_{j-1}, \varphi^i_j)$ \label{alg:smc-reweight-1}
  \State $w^i_j \gets w^i_{j-1} \label{alg:smc-reweight-2}
    \displaystyle\left[
      \frac
      {P_{\mathbf{t}_{1:j}} (k^i, \varphi^i, \mathbf{y}_{1:j})}
      {P_{\mathbf{t}_{1:j-1}} (k^i, \varphi^i, \mathbf{y}_{1:j-1})}
    \right]$
  \If{$j < T \mbox{\bfseries\ and } \textsc{Resample?}(w^{1:M}_j)$} \label{alg:smc-resample-1}
      \State $(\iota_1, \dots, \iota_M) \gets \textsc{Resample}(w^{1:M}_j)$
      \State $(k^i_j, \varphi^i_j) \gets (k^{\iota_i}_{j-1}, \varphi^{\iota_i}_{j-1})$
      \State $w^i_j \gets (w^1_j + \dots + w^M_j) / M$
  \EndIf \label{alg:smc-resample-2}
  \For{$u = 1, \dots, N_{\rm rejuv}$} \label{alg:smc-rejuv-1}
    \State $
      \begin{aligned}[t]
        (k^i_j, \varphi^i_j) &\sim \textsc{InvolutiveMCMC}( \\[-5pt]
          &\quad (k, \varphi) \mapsto P_{\mathbf{t}_{1:j}} (k, \varphi, \mathbf{y}_{1:j}; (k^i_j, \varphi^i_j)))\\[-5pt]
          &\quad\Comment{\footnotesize refer to \cref{sec:smc-imcmc}}
      \end{aligned}$
    \State $\begin{aligned}[t]
      \varphi^i_j &\sim \textsc{HamiltonianMonteCarlo}( \\[-5pt]
          &\quad  \varphi \mapsto P_{\mathbf{t}_{1:j}} (k^i_j, \varphi, \mathbf{y}_{1:j}); \varphi^i_j) \\[-5pt]
          &\quad \Comment{\footnotesize refer to \citet{neal2011}}
          \end{aligned}$
      \label{alg:smc-rejuv-hmc}
  \EndFor \label{alg:smc-rejuv-2}
\EndFor
\end{algorithmic}
\end{algorithm}

\subsection{Involutive MCMC Rejuvenation Moves}
\label{sec:smc-imcmc}
\egroup

The \textit{rejuvenate} step
(\crefrange{alg:smc-rejuv-1}{alg:smc-rejuv-2} of \cref{alg:smc})
evolves the particle $(k^i_{j}, w^i_{j})$ through an MCMC kernel
$K_j$ that adapts the structure and parameters $(k, \varphi)$ to the latest
batch $(\mathbf{t}_j, \mathbf{y}_j)$ of observations.
Rejuvenation is an SMC
step~\labelcrefrange{eq:smc-step-1}{eq:smc-step-2} where the target
distributions are equal, the backward kernel is the time reversal of
$K_j$, and the weight is unchanged.
Since $K_j$ traverses a transdimensional
space~\labelcref{eq:target-support} over symbolic covariance
expressions, we must perform posterior inference over
tree-shaped data structures, {\`a} la \citet{chipman1998,neal2003}.
We build a new class of structure learning moves based on involutive
MCMC~\citep{neklyudov2020,towner2020imcmc} that generalize the greedy
search operators from \citet{duvenaud2013} and embed them in a
Bayesian inference framework.

\subsubsection{Subtree-Replace}
\label{sec:smc-imcmc-subtree-replace}

Starting at $(k, \theta, \eta)$, the \textsc{Subtree-Replace} move
(\cref{fig:imcmc-subtree-replace})
replaces a randomly selected subtree of $(k, \theta)$ with a fresh subtree and
updates the noise $\eta$ using a Gibbs-type move.
More specifically, the proposal $q_{\rm R}$ simulates a path
$b \defas (b_1, \dots, b_m)$
($m \ge 0$, $b_i \in \set{0\defas \mathrm{left},1 \defas\mathrm{right}}$)
to a node in the abstract syntax-tree representation of $k$.
We then simulate a fresh subtree $(\tilde{k}, \tilde{\theta})$
and obtain a proposed state $(k',\theta')$
by replacing the current subtree $(\hat{k}, \hat{\theta})$
rooted at $b$ with $(\tilde{k}, \tilde{\theta})$.
We define notation for $(\hat{k},\hat\theta)$ and $(k',\theta')$ as
\begin{align}
(\hat{k},\hat{\theta}) \gets (k_b, \theta_b),
\;
(k', \theta') \gets
\otimes_b\left( \ominus_b(k, \theta), (\tilde{k}, \tilde{\theta}) \right)
\label{eq:vaginocele}
\end{align}
where $(\cdot)_b$, $\ominus_b(\cdot)$ and $\otimes_b(\cdot,\cdot)$
are operators for extracting, deleting, and inserting subtrees rooted at $b$, respectively.
To propose a new observation noise $\eta'$ we perform a Gibbs-style
move that leverages the conjugacy of the inverse-gamma prior~\labelcref{eq:GP-prior-3}
over $\eta$ and normal likelihood~\labelcref{eq:GP-prior-5}
of the observations $\mathbf{y}$
for fixed $f(\mathbf{t})$ in~\cref{eq:GP-prior-4}.
More specifically, given $(k',\theta',\eta,\mathbf{y})$, the means
$\bm{\mu}' \defas [\mu'_1, \dots, \mu'_n]$ of the random variables
$[f(t_1), \dots, f(t_n)]$ are known in closed form, which lets us
sample $\eta$ from its conditional distribution
given $(k', \theta', f(\mathbf{t}) = \bm{\mu}', \mathbf{y})$
(proofs in \cref{appx:proofs}):
\begin{flalign}
\eta' \sim \mbox{\footnotesize InverseGamma}(1 + n/2, 1 + \textstyle\sum_{i=1}^n(y_i-\mu'_i)^2/2).
\hspace{-.6cm}
\label{eq:eta-gibbs}
&&
\end{flalign}
Let $u \defas (b, \tilde{k}, \tilde{\theta}, \eta')$ be the variables
sampled by $q_{\rm R}$.
Since \textsc{Subtree-Replace} is its own reversal, the mapping
and inverse mapping $[(k, \theta, \eta), u] \leftrightarrow [(k',\theta',\eta'), u']$
are given by a measure-preserving involution $f_{\rm R}$ defined by
\begin{align*}
f_{\rm R}([(k,\theta,\eta), (b, \tilde{k}, \tilde{\theta}, \eta')])
\defas [(k', \theta', \eta'), (b, \hat{k}, \hat{\theta}, \eta)],
\end{align*}
where \cref{eq:vaginocele} defines $(k',\theta')$ and $(\hat{k}, \hat{\theta})$
from $(k,\theta,b,\tilde{k},\tilde{\theta})$.


\begin{figure}[!t]


\centering
\captionsetup[subfigure]{skip=0pt}

\begin{subfigure}[t]{.425\linewidth}
\caption{\textsc{Subtree-Replace}}
\label{fig:imcmc-subtree-replace}
\centering
\begin{tikzpicture}
\tikzset{subtree/.style={
    regular polygon, regular polygon sides=3,
    draw, inner sep=3pt, yshift=.6mm,
    label={[label distance=0.5mm]below:{#1}}}}
\tikzset{branch/.style={circle, draw}}
\tikzset{every tree node/.style={draw,circle}}
\tikzset{level distance=5mm}
\tikzset{sibling distance=1.5mm}
\tikzset{edge from parent/.append style={-}}

\node[
  name=left-tree,
  outer sep=0pt,
  draw=none,
  inner sep=2.5pt,
  fill=white!95!black,
  label={[name=left-tree-label,label distance=0, inner sep=1pt]above:$(k,\theta)$}] {
  \begin{tikzpicture}
  \Tree
    [.{}
      [.{} \node[subtree]{}; ]
      \edge[-latex,thick,color=black]
        node[auto=left,pos=.4,inner ysep=0pt,inner xsep=4pt]{\scriptsize $b$};
      [.{} \node[subtree={$(\hat{k}, \hat{\theta})$}, pattern=crosshatch]{}; ] ]
  \end{tikzpicture}};

\node[
  name=right-tree,
  outer sep=0pt,
  inner sep=2.5pt,
  draw=none,
  fill=white!95!black,
  right=2 of left-tree.north,
  anchor=north,
  label={[name=right-tree-label,label distance=0, inner sep=1pt]above:$(k',\theta')$}] {
  \begin{tikzpicture}[anchor=center]
  \Tree
    [.{}
      [.{} \node[subtree]{}; ]
      \edge[-latex,thick,color=black]
        node[auto=left,pos=.4,inner ysep=0pt,inner xsep=4pt]{\scriptsize $b$};
      [.{} \node[subtree={$(\tilde{k}, \tilde{\theta})$}, pattern=dots]{}; ] ]
  \end{tikzpicture}};

\draw[-latex]
  (left-tree-label.north)
  to [bend left]
  node[pos=.5,anchor=south,font=\footnotesize]{
    $q_{\rm R}\left(b, \tilde{k}, \tilde{\theta}; (k,\theta) \right)$}
  node[pos=.9,anchor=south west, inner ysep=0pt, inner xsep=4pt,font=\footnotesize]{$f_{\rm R}$}
  node[pos=.5,anchor=north,font=\scshape\scriptsize]{Replace}
  (right-tree-label.north);

\draw[-latex]
  ([yshift=-2mm]right-tree.south)
  to [bend left]
  node[pos=.5,anchor=north,font=\footnotesize]{
    $q_{\rm R}\left(b, \hat{k}, \hat{\theta}; (k',\theta')\right)$}
  node[pos=.9,anchor=north east, inner ysep=-2pt,font=\footnotesize]{$f_{\rm R}$}
  node[pos=.5,anchor=south,font=\scshape\scriptsize]{Replace}
  ([yshift=-2mm]left-tree.south);
\end{tikzpicture}
\end{subfigure}\hfill
\begin{subfigure}[t]{.575\linewidth}
\caption{\textsc{Detach-Attach}}
\label{fig:imcmc-detach-attach}
\hfill
\begin{tikzpicture}
\tikzset{level distance=1cm}
\tikzset{subtree/.style={
    regular polygon, regular polygon sides=3,
    draw, inner sep=3pt, yshift=.6mm,
    label={[label distance=0.5mm]below:{#1}}}}
\tikzset{branch/.style={circle, draw, inner sep = 4pt}}
\tikzset{every tree node/.style={draw,circle}}
\tikzset{level distance=5mm}
\tikzset{sibling distance=-1mm}
\tikzset{edge from parent/.append style={-}}

\node[
  name=left-tree,
  outer sep=0pt,
  draw=none,
  inner sep=2.5pt,
  fill=white!95!black,
  label={[name=left-tree-label,label distance=0, inner sep=1pt]above:$(k,\theta)$}] {
  \begin{tikzpicture}
  \Tree
    [.{}
      \edge[-latex,thick,color=black]
        node[auto=right,pos=.4,inner ysep=0pt,inner xsep=4pt]{\scriptsize $a$};
      [.\node[name=salman]{};
          [.{} \node[name=mansour,subtree={$(\hat{k}, \hat{\theta})$}]{}; ]
          \edge[-latex,thick,color=black]
            node[auto=left,pos=.4,inner sep=1pt]{\scriptsize $b$};
          [.{} \node[subtree={$(\tilde{k}, \tilde{\theta})$}, pattern=crosshatch]{}; ] ]
      [.{} \node[subtree]{}; ] ]
  \draw[dotted, thick]
    (salman.north)
    to [bend right] ([xshift=-2mm,yshift=-2mm]mansour.west)
    to ([xshift=2mm,yshift=-2mm]mansour.east)
    to [bend right] ([xshift=2mm,yshift=0mm]mansour.north)
    to [bend left] (salman.south)
    ;
  \end{tikzpicture}};

\node[
  name=right-tree,
  outer sep=0pt,
  draw=none,
  inner sep=2.5pt,
  fill=white!95!black,
  right=2.2 of left-tree.north,
  anchor=north,
  label={[name=right-tree-label,label distance=0, inner sep=1pt]above:$(k',\theta')$}] {
  \begin{tikzpicture}[anchor=center]
  \Tree
    [.{}
      \edge[-latex,thick,color=black]
        node[auto=right,pos=.4,inner ysep=0pt,inner xsep=4pt]{\scriptsize $a$};
      [.{} \node[subtree={$(\tilde{k}, \tilde{\theta})$},pattern=crosshatch]{}; ]
      [.{} \node[subtree]{}; ] ]
  \end{tikzpicture}};

\draw[-latex]
  (left-tree-label.north)
  to [bend left]
  node[pos=.5,anchor=south,font=\footnotesize]{$q_{\rm D}\left(a,b; (k,\theta)\right)$}
  node[pos=.5,anchor=north,font=\scshape\scriptsize]{Detach}
  node[pos=.9,anchor=south west, inner ysep=0pt, inner xsep=4pt,font=\footnotesize]{$f_{\rm D}$}
  (right-tree-label.north);

\draw[-latex]
  (right-tree.south)
  to [bend left]
  node[pos=.5,anchor=north,draw=none,xshift=-5mm,yshift=-2mm,font=\footnotesize]{
    $q_{\rm A}\left(a,b,\hat{k}, \hat{\theta}; (k',\theta') \right)$}
  node[pos=.5,anchor=west,,xshift=8pt,font=\scshape\scriptsize]{Attach}
  node[pos=.9,anchor=north east, inner ysep=0.25pt,font=\footnotesize]{$f_{\rm A}$}
  (left-tree.south);
\end{tikzpicture}
\end{subfigure}
\vspace{-10pt}
\caption{Involutive MCMC rejuvenation moves over the structure $k$ and parameters
$\theta$ of a Gaussian process covariance function $k_\theta$.}
\label{fig:imcmc}
\end{figure}

\begin{proposition}
\label{prop:subtree-replace}
If the proposal $q_{\rm R}$ samples the path $b$ uniformly and
$(\tilde{k},\tilde{\theta})$ from the conditional prior
given $({\ominus}_b(k, \theta))$, then the involutive MCMC
acceptance probability $\alpha_{\rm R} = \min(1, r_{\rm R})$, where
\begin{align}
r_{\rm R} \defas
\frac{P(\eta')}{P(\eta)}
\frac
  {P_{\mathbf{t}}(\mathbf{y} \mid k',\theta',\eta')}
  {P_{\mathbf{t}}(\mathbf{y} \mid k,\theta,\eta)}
\frac{1/\abs{k'}}{1/\abs{k}}
\frac
  {P_{\mathbf{t}}(\eta \mid \bm{\mu}, \mathbf{y})}
  { P_{\mathbf{t}}(\eta' \mid \bm{\mu}', \mathbf{y})}.
\label{eq:subtree-replace-accept}
\end{align}
\end{proposition}

\begin{remark*}
Even though the proposal~\labelcref{eq:eta-gibbs} for $\eta'$
resembles a Gibbs move, the ratio $r_{\rm R}$ includes terms for
$\eta'$ because it is sampled \textit{before} deciding whether to
accept $(k', \theta')$.
\end{remark*}

While the \textsc{Subtree-Replace} proposal is very flexible
and, provided that $b$ is root with positive probability,
ensures irreducibility of the chain, it has a problematic limitation.
Consider a proposed move $(k, \theta) \to (k', \theta')$ with
\begin{align}
\underbrace{\textsc{Linear}_{\theta_1}}_{{(k,\theta)}} \to
  \underbrace{
    \textsc{(}
    \textsc{Linear}_{\theta'_1}
    \textsc{ + }
    \textsc{Periodic}_{\theta'_2,\theta'_3}
    \textsc{)}
  }_{(k',\theta')}
\label{eq:linear-to-per}
\end{align}
which may arise in the transition from \cref{fig:real-time-25}
to \cref{fig:real-time-50}.
The only way for \textsc{Subtree-Replace} to perform such a move
is to select the root and resimulate the entire expression, which
\begin{enumerate*}[label=(\roman*)]
\item has a positive
probability of not including $\textsc{Linear}$; and
\item discards the inferred parameter $\theta_1$ in the current state.
\end{enumerate*}
The next move type is designed to address this limitation.

\subsubsection{Detach-Attach}
\label{sec:smc-imcmc-detach-attach}

Starting at $(k, \theta, \eta)$, the \textsc{Detach-Attach} move
(\cref{fig:imcmc-detach-attach})
is composed of a pair of submoves that invert one another.
\begin{itemize}[wide=0pt,itemsep=0pt,topsep=0pt]

\item \textsc{Detach}: Replaces the subtree $(k_a, \theta_a)$ rooted at
$a$ with the subtree $(\tilde{k}, \tilde{\theta})$ rooted at $(a,b)$;
so the proposed state is $(k', \theta') \gets \otimes_a (\ominus_a(k,\theta), (\tilde{k}, \tilde{\theta}))$.
In \cref{fig:imcmc-detach-attach}, the dotted region shows the
discarded fragment $(\hat{k}, \hat{\theta})$ of $(k_a, \theta_a)$,
which will be the ``scaffold'' that must be simulated
in the corresponding \textsc{Attach} direction to reverse this move.

\item \textsc{Attach}: Replaces the subtree $(\tilde{k}, \tilde{\theta})$
rooted at $a$ with a new subtree $(k_a, \theta_a)$
that itself embeds $(\tilde{k}, \tilde{\theta})$ as a subtree rooted at $b$.
The remaining components of $(k_a, \theta_a)$
other than $(\tilde{k}, \tilde{\theta})$, i.e., the scaffold, are precisely
the discarded fragment $(\hat{k}, \hat{\theta})$ from the corresponding \textsc{Detach} move.
The proposed state is
$(k',\theta') \gets \otimes_a( \ominus_a(k, \theta), (k_a, \theta_a))$
\end{itemize}

Whereas \textsc{Subtree-Replace} is self-inverting, the \textsc{Detach-Attach}
move is associated with two proposals $q_{\rm D}$ and $q_{\rm A}$
and two maps $f_{\rm D}$ and $f_{\rm A}$ that invert one another.
The $\textsc{Detach}$ direction is chosen with probability $\xi \in (0,1)$.
Proposal $q_{\rm D}$ simulates a path $(a,b)$
whereas $q_{\rm A}$ simulates both a path $(a,b)$ and a scaffold $(\hat{k}, \hat{\theta})$.
Moreover
\begin{align}
f_{\rm D}([(k, \theta, \eta), (a, b)]) \defas [(k', \theta', \eta'), (a, b, \hat{k}, \hat{\theta},\eta)],  \label{eq:inv-detach} \\
f_{\rm A}([(k, \theta, \eta), (a, b, \hat{k}, \hat{\theta})]) \defas [(k', \theta', \eta'), (a, b, \eta)].  \label{eq:inv-attach}
\end{align}

\begin{proposition}
\label{prop:attach-detach}
If $q_{\rm D}$ and $q_{\rm A}$ sample paths $a$ uniformly
and $q_{\rm A}$ samples
scaffold $(\hat{k}, \hat{\theta})$ from the conditional prior
given $[\ominus_a (k, \theta), b, (\tilde{k}, \tilde{\theta})]$, then the acceptance ratios
$r_{\rm D} = r_{\rm R}
  \cdot (1-\xi)/\xi
  \cdot q_{\rm A}(b\,{\mid}\,a; k',\theta')/q_{\rm D}(b\,{\mid}\,a; k, \theta)$
and
$r_{\rm A} = r_{\rm R}^2 r_{\rm D}^{-1}$.
\end{proposition}

\begin{figure}[!t]
\includegraphics[width=.33\linewidth]{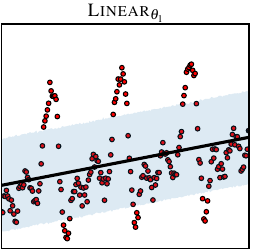}%
\includegraphics[width=.33\linewidth]{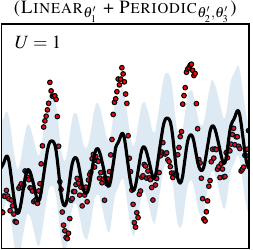}%
\includegraphics[width=.33\linewidth]{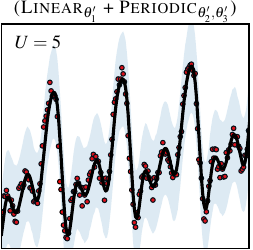}
\includegraphics[width=\linewidth]{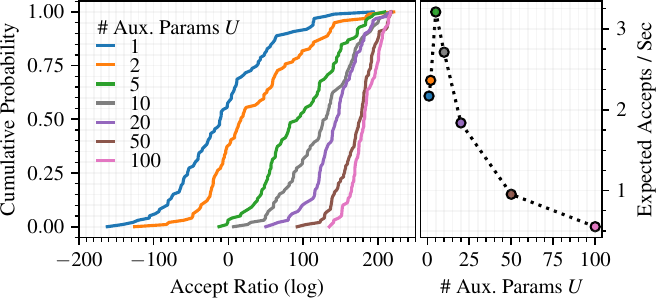}
\caption{
Top row: Structure proposal~\labelcref{eq:linear-to-per} using
$U\,{\in}\,\set{1,5}$ auxiliary parameters.
Bottom row: Distribution of log acceptance ratios~\labelcref{eq:portability}
and expected acceptances per second for various $U$.}
\label{fig:sir}
\end{figure}

\subsubsection{Improving Acceptance Rates via Pseudo-Marginal Parameter Proposals}
\label{sec:smc-imcmc-pseudo-marginal}

\begin{figure*}[!t]
\centering
\refstepcounter{subfigure}\label{fig:runtime-airline}
\refstepcounter{subfigure}\label{fig:runtime-mauna}
\refstepcounter{subfigure}\label{fig:runtime-gasoline}
\refstepcounter{subfigure}\label{fig:runtime-radio}
\refstepcounter{subfigure}\label{fig:runtime-housing}
\refstepcounter{subfigure}\label{fig:runtime-australia}
\refstepcounter{subfigure}\label{fig:runtime-internet}
\refstepcounter{subfigure}\label{fig:runtime-fatalities}
\includegraphics[width=\linewidth]{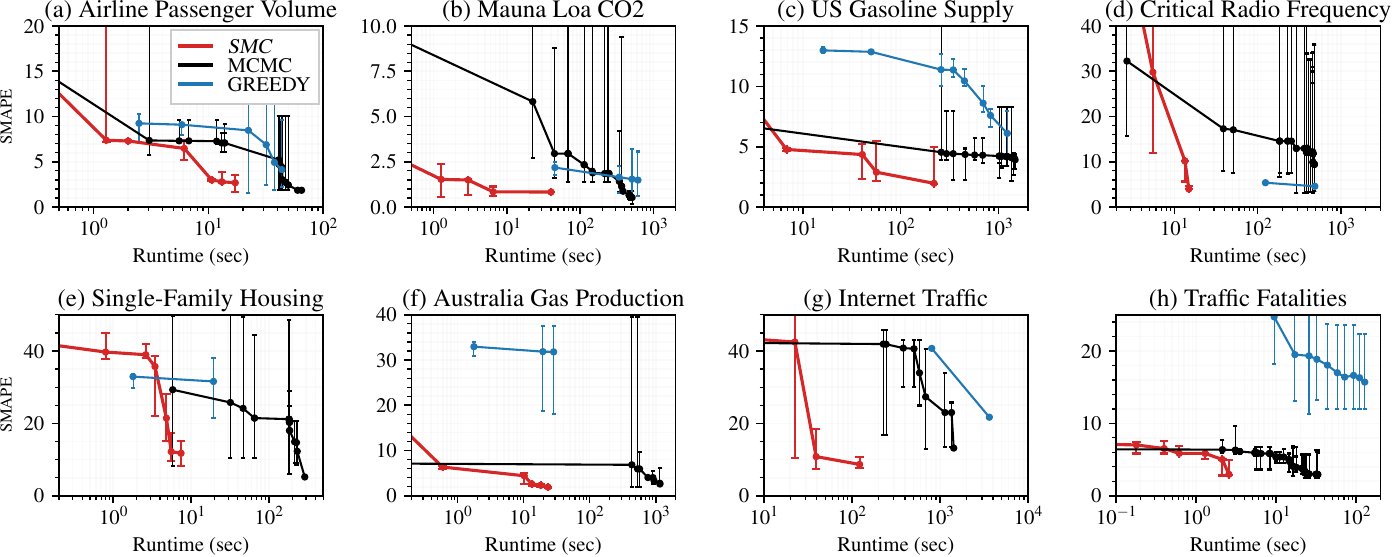}
\caption{Runtime (x-axes, log scale) vs.~forecast error
(y-axes, SMAPE over 18-step forecast horizon) for eight
datasets in the Time \mbox{Series} Data Library~\citep{hyndman2018}
using three different structure discovery algorithms for
temporal Gaussian processes:
SMC sampling (\cref{alg:smc}),
MCMC sampling \citep{saad2019}
and GREEDY search \citep{duvenaud2013}.
}
\label{fig:runtime}
\end{figure*}

It is well known that transdimensional MCMC
algorithms~\citep{green1995} often suffer from high rejection rates
for moves $(k, \theta) \to (k', \theta')$ that change the
structure~\citep{brooks2003,karagiannis2014}.
In particular, even though $k'$ may model the data $\mathbf{y}$ well
using some ideal parameter $\theta'_*$, the transition is likely
rejected if the
proposed parameter $\theta'$ is such that $\mathbf{y}$ has low
density given $(k', \theta')$.
The first two panels in the top row of \cref{fig:sir} illustrate
this idea for the move~\labelcref{eq:linear-to-per},
where the proposed parameters $\theta'_2, \theta'_3$ in the
$\textsc{Periodic}$ subexpression have incorrect frequency and lengthscale,
leading to a poor fit.
We address this problem by developing an auxiliary-variable method to improve
$\theta'$ during the transdimensional proposal.

For the move $(k, \theta) \to (k', \theta')$, we partition
$\theta = (\theta_S, \theta_D)$ and $\theta' = (\theta'_S,
\theta'_F)$, where $S$ indexes the ``shared'' parameters that exist in both
$\theta$ and $\theta'$ (which may have different values), $D$ indexes
the ``discarded'' parameters from $\theta$ that do not exist in
$\theta'$, and $F$ indexes the ``fresh'' parameters in $\theta'$ that
do not exist in $\theta$.
The proposal simulates $U > 0$ i.i.d.~auxiliary parameters
$\set{(\theta'^u_S, \theta'^u_F)}_{u=1}^U$ using log-normal
random walk proposals (for $S$) and prior proposals (for $F$)
and selects one $\tilde{u} \in [U]$ to serve as the selected
parameter $\theta'$:
\begin{align}
\theta'^u_s &\sim \mbox{\footnotesize{LogNormal}}(\theta_{s,\delta}, \delta) && (u=1\dots U; s \in S)
\label{eq:sir-11} \\
\theta'^u_f &\sim \mbox{\footnotesize{LogNormal}}(0, 1) && (u=1\dots U; f \in F)
\label{eq:sir-12}\\
\tilde{u} &\sim \mbox{\footnotesize{Discrete}}
\left(
  \omega(\theta'^1_{S,F}), \dots, \omega(\theta'^U_{S,F})
\right), \span\span
\label{eq:sir-13}
\end{align}
where $\theta_{s,\delta} \defas \ln(\theta_s)-\delta/2$ ($\delta \ge 0$)
and $\omega(\cdot)$ assigns a nonnegative weight
to each $\theta'^{u}_{S,F}$.
The optimal choice of $\omega$
that minimizes the variance of the estimate of the marginal
likelihood $P_{\mathbf{t}}(k' \mid \mathbf{y})$ is precisely the importance weight
\begin{align}
\omega(\theta'^u_{S,F}) =
  \frac{P_{\mathbf{t}}(k', \theta'^u_{S,F}, \eta, \mathbf{y})}
  {
    \prod_{s \in S} q(\theta'^u_s; \theta_s)
    \prod_{s \in S} q(\theta'^u_f)
  },
\label{eq:sir-optimal}
\end{align}
although the general theory we develop allows any non-negative
$\omega$, which may be faster to compute than \cref{eq:sir-optimal}.
As part of the forward proposal $(k,\theta) \to (k', \theta')$,
we also simulate $U-1$ auxiliary ``reverse'' parameters,
\begin{align}
\theta^{u}_{s} &\sim \mbox{\footnotesize LogNormal}({\theta'^{\tilde{u}}_{s,\delta}}, \delta) && \hspace{-.2cm} (u=1\dots U-1; s \in S)
\label{eq:sir-21} \\
\theta^{u}_{d} &\sim \mbox{\footnotesize LogNormal}(0, 1) && \hspace{-.25cm} (u=1\dots U-1; d \in D)
\label{eq:sir-22} \\
\hat{u} &\sim \mbox{\footnotesize Uniform}(1, \dots, U). \span\span
\label{eq:sir-23}
\end{align}
The role of these parameters is described in the next
proposition, which presents a proposal distribution, involution, and
acceptance ratio for an involutive MCMC scheme using the auxiliary-variable
proposals~\labelcref{eq:sir-11,eq:sir-12,eq:sir-13,eq:sir-21,eq:sir-22,eq:sir-23}.

\begin{proposition}
\label{prop:subtree-replace-pseudo}
Consider the \textsc{Subtree-Replace} move using an extended proposal $q_{\rm R}^U$
that samples auxiliary variables~\labelcref{eq:sir-11,eq:sir-12,eq:sir-13,eq:sir-21,eq:sir-22,eq:sir-23},
equipped with the involution
\begin{align}
&\begin{aligned}[b]
&f^U_{\mathrm{R}}([
  (k, \theta, \eta),
  (b, \tilde{k}, \theta'^{1:U}_{S,F}, \tilde{u}, \theta^{1:U-1}_{S,D}, \hat{u}, \eta')
    ])\\
&\begin{aligned}[t]
  \quad =[(k', \theta', \eta'),
  (b, \hat{k},
    &(\theta^{1:\hat{u}-1}_{S,D},\theta_{S,D},\theta^{\hat{u}:U-1}_{S,D}), \hat{u}, \\
    &(\theta'^{1:\tilde{u}-1}_{S,F},\theta'^{\tilde{u}+1:U}_{S,F}), \tilde{u},
    \eta)]
    \end{aligned}
\end{aligned}
\label{eq:subtree-replace-inv1}
\\
&\mbox{where }\begin{aligned}
(k', \theta') \defas
\otimes_b\left(
  \ominus_b(k, (\theta'^{\tilde{u}}_S, \theta_D)),
    (\tilde{k}, {\theta'^{\tilde{u}}_F})
    \right).
\end{aligned}
\label{eq:subtree-replace-inv2}
\end{align}
If $q^U_{\mathrm{R}}$ samples the path $b$ uniformly
and samples $\tilde{k}$ from
the conditional prior given $(\ominus_b\ k)$, then involutive MCMC with acceptance
probability $\alpha^U_{\rm R} \defas \min(1, \tilde{r}^U_{\mathrm{R}})$
defines an irreducible, aperiodic Markov chain with stationary
distribution $P_{\mathbf{t}}(k,\varphi\,{\mid}\,\mathbf{y})$,
where
$\tilde{r}^U_{\mathrm{R}}$ is defined in~\cref{eq:portability} of
\cref{appx:proofs}.
\end{proposition}

Analogous results hold for the \textsc{Attach-Detach} move.

\paragraph{Example}
Consider the move in \cref{eq:linear-to-per}.
The top row of \cref{fig:sir} shows how the current \textsc{Linear}
model $(k,\theta)$ and two proposed models $(k',\theta')$
fit the data, with $U=1,5$ auxiliary
parameters for sampling $\theta'$.
The second row shows the empirical distribution of acceptance
ratios $\tilde{r}^U_{\mathrm{R}}$ for various $U$.
Since $k'$ is marginally orders of magnitude more likely for the
observed data than $k$, increasing $U$
leads to $\min(1, \tilde{r}^U_{\rm R}) = \alpha^U_{\rm R} \approx 1$ with
overwhelming probability.
Letting $t_{U}$ denote the number of seconds needed to compute
$\tilde{r}^U_{\rm R}$ and
$N^{\mathrm{accept}}_{U} = \alpha^U_{\rm R}/t_{U}$
the number of accepts per second,
\cref{fig:sir} shows that increasing $U$ eventually leads to
diminishing returns of
$\mathbb{E}\left[N^{\mathrm{accept}}_{U}\right]$, because
$\alpha_U$ is essentially 1 for $U \ge 5$.
While the optimal value of $U$ is difficult to quantify theoretically
for arbitrary moves $(k, \theta) \to (k', \theta')$, one possible
choice is to set $U \propto \abs{\theta'} \equiv d(k')$.

\section{Evaluation}
\label{sec:evaluation}

We implemented\footnote{Available online at \url{https://github.com/fsaad/AutoGP.jl}} the
time series structure discovery method described in \cref{sec:smc}
using the Gen probabilistic programming system~\citep{towner2019gen}
and evaluated its performance against multiple automated forecasting
methods.
\Cref{sec:evaluation-runtime} compares runtime versus accuracy
profiles to two previous algorithms for Gaussian processes structure
learning and \cref{sec:evaluation-m3} presents runtime and accuracy
results on challenging econometric forecasting benchmarks.
All the experiments were conducted on a Google Cloud n2d-standard-48 instance
(server specs: AMD EPYC\textsuperscript{\texttrademark} 7B12 48vCPU
@2.25GHz, 192 GB RAM).

\begin{figure*}[t]
\includegraphics[width=\linewidth]{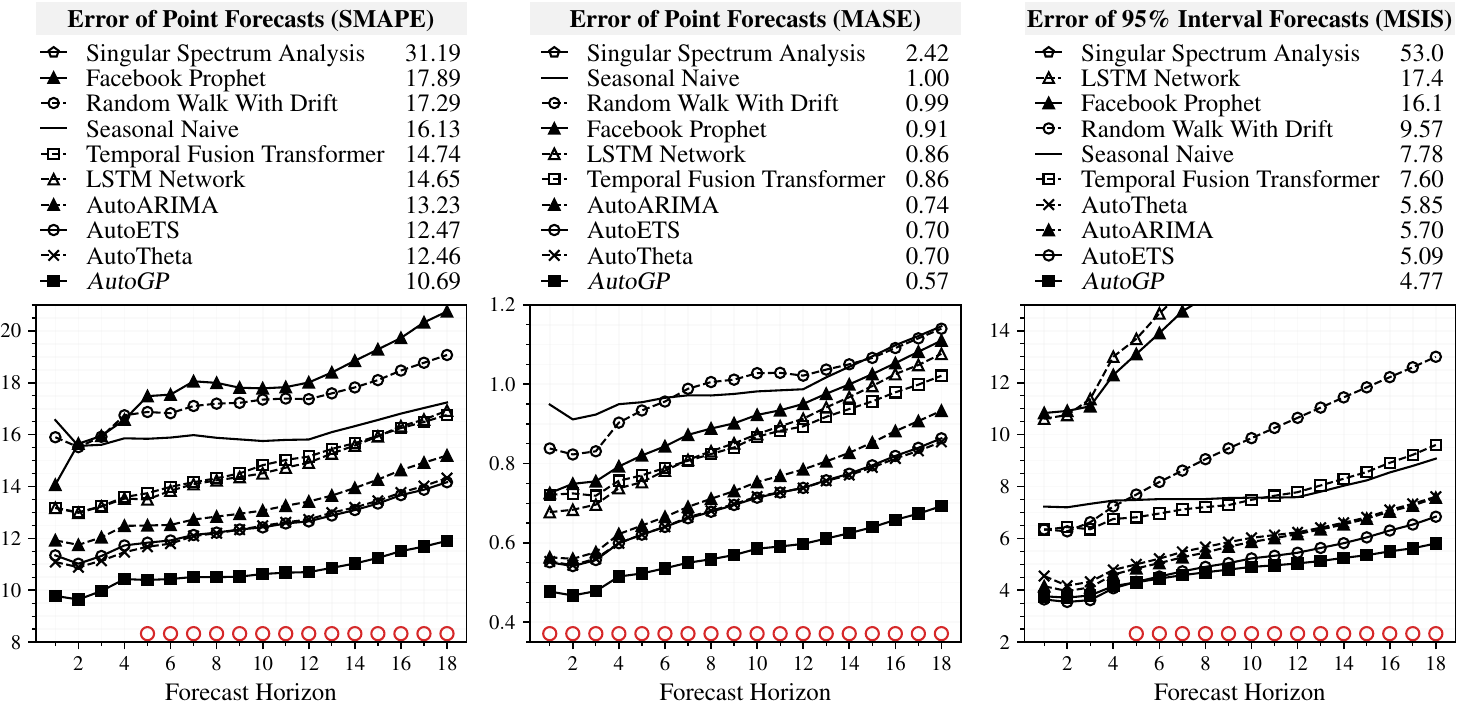}
\caption{Error of point forecasts (SMAPE and MASE) and 95\% interval forecasts
(MSIS) using our method (labeled \AutoGP{}) and several baselines over 18
forecast horizons. Each dot at a given horizon shows the mean error
(SMAPE, MASE, MSIS) across 1,428 forecasting problems in
the M3 monthly econometric time series~\citep{makridakis2000}.
Red circles indicate horizons where errors of \AutoGP{} are statistically
significantly less than errors of the next-best method (Mann-Whitney U
test, $p=0.05$ with Bonferroni correction for 54 tests). Legends show
in sorted order the overall error of each method, which is the mean
error across all 18 horizons.}
\label{fig:m3-eval}
\end{figure*}

\subsection{Runtime Comparisons to MCMC \& Greedy Search}
\label{sec:evaluation-runtime}

\Cref{fig:runtime} shows runtime vs.~forecasting accuracy profiles for
eight TSDL datasets~\citep{hyndman2018} using our SMC
sampler (\cref{alg:smc}), MCMC sampling \citep{saad2019}, and GREEDY
search~\citep{duvenaud2013}.
As in~\citet{duvenaud2013}, for GREEDY the maximum search depth is 10, random
restarts are used during parameter optimization, and structures
are scored using the BIC.
As our server has $M=48$ parallel threads, we used $M$ particles for
SMC, ran $M$ parallel chains for MCMC, and scored $M$ structures in
parallel for greedy search.

Greedy search delivers forecasts of comparable accuracy to SMC and MCMC
in three datasets
(\cref{fig:runtime-airline,fig:runtime-mauna,fig:runtime-radio})
and lower accuracy in five datasets
(\cref{fig:runtime-gasoline,fig:runtime-housing,fig:runtime-australia,fig:runtime-internet,fig:runtime-fatalities}).
In \cref{fig:runtime-australia,fig:runtime-fatalities},
the errors from greedy search after 10 and 100 seconds have high
variance (due to random restarts) and are over 2x higher than those
obtained from SMC in 1 second.
A main distinction between greedy search operators and MCMC
rejuvenation operators is that the former can never decrease the size
of the kernel structure at given step, whereas the latter makes
stochastic accept/reject choices using reversible moves that leave the
posterior invariant.

The error distributions using SMC and MCMC typically converge to
similar values, although the MCMC chains exhibit higher variance
and can require between 10x
(\cref{fig:runtime-internet,fig:runtime-fatalities}) to 100x
(\cref{fig:runtime-radio,fig:runtime-australia}) more wall-clock time.
It is helpful to compare SMC and MCMC using the same
overall rejuvenation budget.
Consider $n$ observations and an SMC scheme with a logarithmic
annealing schedule $\mathbf{t}_{1:2}, \mathbf{t}_{1:4}, \dots, \mathbf{t}_{1:n}$
(i.e., $\log{n}$ steps) and $r$ rejuvenation moves per step.
As the rejuvenation cost with $n$ observations is $O(n^3)$ for a dense
Gaussian process, the total cost of rejuvenation within an SMC particle is
$\sum_{i=0}^{\log{n}} r(n/2^i)^3 = O(rn^3)$.
An MCMC chain with the same budget performs $O(r\log(n) n^3)$ work.
We thus expect SMC to deliver speedups when structures can be inferred
using $n_*\,{\ll}\,n$ data points relatively early in the annealing
schedule.

\begin{figure*}[!t]
\captionsetup[subfigure]{skip=0pt}
\begin{subfigure}[t]{.325\linewidth}
\caption{Shipments of Paints/Laquer (M3-N2015)}
\label{fig:m3-explore-paints}
\includegraphics[width=\linewidth]{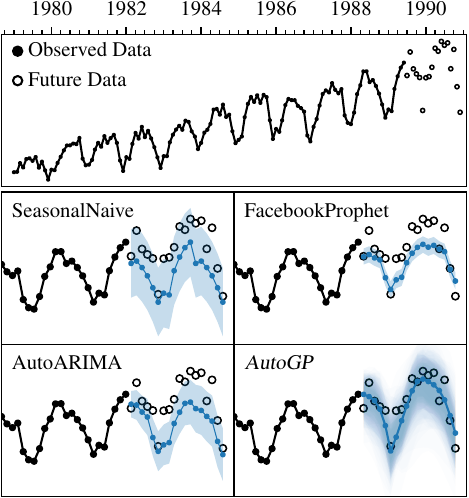}
\end{subfigure}\hfill
\begin{subfigure}[t]{.325\linewidth}
\caption{Telecommunications Vol. (M3-N2803)}
\label{fig:m3-explore-telecom}
\includegraphics[width=\linewidth]{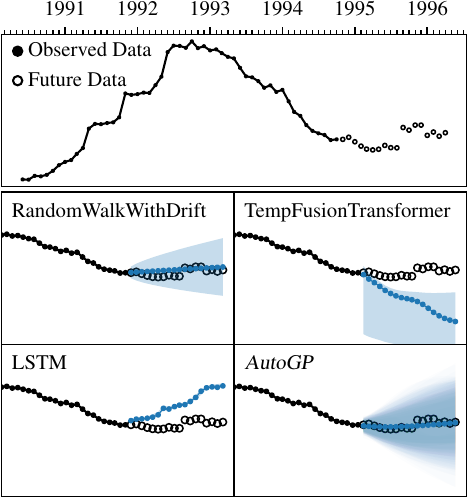}
\end{subfigure}\hfill
\begin{subfigure}[t]{.325\linewidth}
\caption{Civilian Labor Force (M3-N2753)}
\label{fig:m3-explore-labor}
\includegraphics[width=\linewidth]{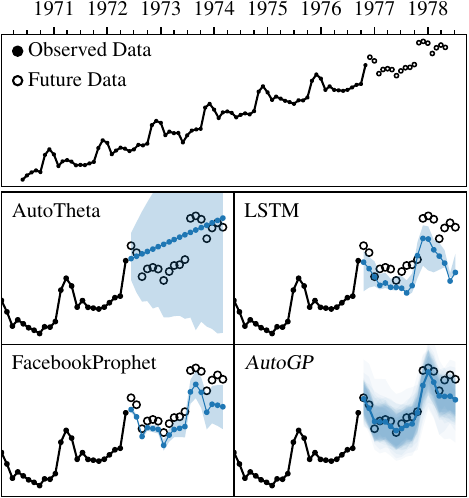}
\end{subfigure}
\caption{Forecasts on three example time series in the
M3 data using our method (labeled \textsl{AutoGP}) and baselines from the evaluation in
\cref{fig:m3-eval}.
Observed data is shown in black;
future data is shown in white;
and point forecasts and 95\% interval forecasts are shown in blue.
}
\label{fig:m3-explore}
\end{figure*}

\subsection{Forecasting Accuracy on Econometric Data}
\label{sec:evaluation-m3}

\paragraph{Motivation} The next evaluation builds on the study
of~\citet{makridakis2018} (which itself builds on the earlier study of
\citet{ahmed2010}) comparing the accuracy of popular machine learning
forecasting methods to statistical baselines in a benchmark of
over 1,428 monthly time series from the M3
dataset~\citep{makridakis2000}.
This previous study found that statistical methods consistently
outperform more complex machine learning methods across all 18
forecasting horizons.
Follow-on work by \citet{cerqueira2022} showed that increasing the
number of observations improves the performance of machine learning
methods to some extent, but they still do not outperform statistical
baselines on the 18-step forecast horizon.\footnote{The more recent
study of \citet{makridakis2023} found that deep learning ensembles can
perform competitively to statistical ensembles on the M3 benchmark,
but they require several months of training and hyperparameter tuning.
In particular, the authors of the study write that: \textit{``the time
spent for tuning the four deep learning methods was 3,193, 512, 507
and 3,077 hours for DeepAR, Feed-Forward, Transformer, and WaveNet,
respectively''}, which renders these methods highly impractical for
the typical user.}

We benchmarked our time series method on the M3 monthly data
and compared the results to top-performing baselines from
\citet{makridakis2018} and other popular methods that also deliver
probabilistic forecasts, including:
  AutoARIMA~\citep{hyndman2008},
  AutoETS~\citep{hyndman2008ets},
  AutoTheta~\citep{fiorucci2016},
  Facebook Prophet~\citep{taylor2018},
  Singular Spectrum Analysis~\citep{golyandina2018},
  LSTM~\citep{lindemann2021},
  and Temporal Fusion Transformer~\citep{lim2021}.
Refer to \cref{appx:evaluation} for additional experimental details.

\paragraph{Accuracy} \Cref{fig:m3-eval} shows the forecasting accuracy results.
As in \citet{makridakis2018}, point forecasts are evaluated using the
SMAPE and MASE metrics and 95\% interval forecasts are evaluated
using the MSIS metric (refer to \crefrange{eq:smape}{eq:msis} in \cref{appx:evaluation}).
Since fitting procedures for most forecasting methods are themselves
stochastic, we report the lowest error across 10 different random
seeds, which lets us better quantify the predictive capacity of each
baseline independently of stochasticity in the fitting process.
Our method (named \AutoGP{}, analogously to e.g., AutoARIMA)
delivers the lowest forecast errors across all three metrics, which are
statistically significant in horizons 5--18 for SMAPE, 1--18 for
MASE, and 5--18 for MSIS.

\paragraph{Qualitative Plots}
To gain more insight into the qualitative differences between our
method and the baselines, we investigate forecasts on three example M3
time series in more detail.
In \cref{fig:m3-explore-paints}, the baselines capture the seasonal
variation, however Seasonal Naive and AutoARIMA produce forecasts
that are too low and, in the latter's case, with
insufficient uncertainty.
The forecast intervals from Facebook Prophet are also too narrow, a
property that appears in several datasets (e.g.,
\cref{fig:m3-explore-labor}) and is also observed
in~\citet[Section~12.2]{hyndman2021}.
In \cref{fig:m3-explore-telecom}, Random Walk With Drift
accurately explains the data and this structure is emulated by our
method, whereas LSTM and Temporal Fusion Transformer both capture
incorrect dynamics.
\Cref{fig:m3-explore-labor} shows an example of how AutoTheta, a
top-performing and robust statistical baseline on M3, often
learns conservative models that underfit.
LSTM and Facebook Prophet capture the seasonal structure but their
point forecasts are too low and interval forecasts poorly calibrated.
Our method also under-predicts the last three data points in
\cref{fig:m3-explore-labor} but they fall within the 95\%
prediction interval.

\paragraph{Runtime}
\Cref{tbl:m3-runtime} shows runtime statistics per M3 dataset of each
method.
The average runtime of our method (\AutoGP{}, 34.68 sec) lies
between that of AutoETS (21.09 sec) and that of LSTM (36.71 sec).
Its higher variance (8.59 sec) as compared to the baselines is due to
the fact that the complexity of the learned Gaussian process
structures (which take values in an unbounded space,
c.f.~\crefrange{eq:pcfg-1}{eq:pcfg-3}) depends on the
complexity of the underlying temporal patterns in the observed data,
whereas the baselines typically optimize over a fixed model and parameter
space, making their runtime more stable across different datasets.

The AutoARIMA, AutoTheta, and AutoETS baselines all
leverage stepwise, enumeration-based model search algorithms that do
not provide a straightforward way for users to control accuracy
as a function of runtime. In contrast, users can control the runtime
and accuracy trade-offs for \AutoGP{} (as in \cref{fig:runtime}, for
example) by modifying inference hyperparameters of \cref{alg:smc} such
as the number $M$ of SMC particles, number $N_{\mathrm{rejuv}}$ of
involutive MCMC rejuvenation steps, or number $T$ of annealing steps.

\begin{table}[!b]
\captionsetup{belowskip=-20pt}
\caption{Mean and standard deviation of wall-clock runtime
per dataset for various forecasting baselines in the M3 evaluation.}
\label{tbl:m3-runtime}
\footnotesize
\begin{tabular*}{\linewidth}{@{\extracolsep{\fill}}lrr}
\toprule
{}                          & \multicolumn{2}{l}{\textnormal{Runtime / Dataset} (sec)} \\
{}                          & Mean   & Std \\
\midrule
Facebook Prophet            & 0.87   & 0.41 \\
Historic Average            & 4.02   & 1.07 \\
RandomWalk With Drift       & 0.01   & 1.25 \\
Naive                       & 0.01   & 1.40 \\
Seasonal Naive              & 5.20   & 1.64 \\
AutoARIMA                   & 12.79  & 2.31 \\
AutoTheta                   & 20.39  & 2.61 \\
AutoETS                     & 21.09  & 3.26 \\
\AutoGP{}                   & 34.68  & 8.59 \\
LSTM                        & 36.71  & 0.34 \\
Temporal Fusion Transformer & 414.80 & 0.34 \\
\bottomrule
\end{tabular*}
\end{table}

\section{Related Work}
\label{sec:related}

\paragraph{Gaussian Process Modeling}

Gaussian processes are widely used models for time series
data~\citep{roberts2013}.
Our work builds on the automated covariance kernel discovery approach
from \citet{duvenaud2013} by introducing a new structure learning
algorithm based on sequential Monte Carlo sampling in a Bayesian
model over covariance expressions, numeric parameters, and data.
Using this new algorithm, this work presents a quantitative evaluation
that goes beyond previous work in the automated Gaussian process model
discovery literature in terms of the number of datasets (D=1428) and
baselines (B=9), e.g.,
\citet[][D=5, B=5]{duvenaud2013};
\citet[][D=13, B=7]{lloyd2014};
\citet[][D=0, B=0]{janz2017};
\citet[][D=6, B=0]{kim2018};
\citet[][D=4, B=5]{schaechtle2017}; and
\citet[][D=7, B=5]{saad2019}.
\Citet{berns2022} provide a survey of automated covariance
kernel discovery for Gaussian process regression models with multiple
input dimensions.
While our work focuses specifically on a language
\labelcrefrange{eq:pcfg-1}{eq:pcfg-3} of kernel structures that specify
covariance functions for a single temporal input, it would be fruitful
to apply our SMC structure learning algorithm to more expressive
modeling languages that support multiple inputs.

\paragraph{SMC Structure Learning}
Sequential Monte Carlo learning algorithms have been used for Bayesian
inference over structured latent spaces in a variety of settings,
including
probabilistic graphical models~\citep{hamze2005,yu2021,naesseth2014},
phylogenetic tree models~\citep{wang2015,wang2019,moretti2021},
and Dirichlet process-based models~\citep{obermeyer2014,saad2018trcrpm}.
Our approach in \cref{alg:smc} uses resample-move
SMC~\citep{chopin2002}, where the sequence of target distributions is
obtained via data annealing and involutive MCMC rejuvenation moves
(\cref{fig:imcmc}) are used to adapt the structure given new
observations; which is a specific instantiation of the general
algorithm template in~\citet[Section 3.3.2]{saad2022} for Bayesian
structure learning via SMC.
\Cref{fig:runtime} suggests that applying this SMC algorithm template in
an analogous manner to \cref{alg:smc} may improve upon MCMC algorithms
for structure learning in other settings, such as the
model families in
\citet{adams2010,mansinghka2016,jin2019,cranefield2017,saad2021hirm}.

\paragraph{Transdimensional Auxiliary Variables}
\Cref{sec:smc-imcmc-pseudo-marginal} introduces a
sampling-importance-resampling proposal for parameters in a
transdimensional move, which we formally justify as an involutive
MCMC step \citep{neklyudov2020,towner2019gen} over an extended
state-space with auxiliary variables~\citep{lew2022}.
Previous works have also used auxiliary-variable strategies for transdimensional
MCMC~\citep{brooks2003,alawadhi2004,andrieu2009}---the most
closely related methods are those in \citet{yeh2012,karagiannis2014}.
These two works use a single-particle, multi-step annealed importance
sampler (AIS) for proposing parameters in a transdimensional move;
whereas our method in \cref{sec:smc-imcmc-pseudo-marginal} can be
seen as a multi-particle, single-step AIS proposal.
A natural idea is to combine these approaches by using a
multi-particle, multi-step AIS transdimensional parameter proposal,
which can be formally understood as an involutive
MCMC move with a proposal pair as in \citet[Algorithms 3 and 4]{saad2022eevi}
and an involution function that generalizes the one
in~\crefrange{eq:subtree-replace-inv1}{eq:subtree-replace-inv2}.

\section{Conclusion}
\label{sec:conclusion}

We have presented a new approach for automatically discovering models
of time series data.
Our method leverages sequential Monte Carlo inference (\cref{alg:smc})
to effectively infer the posterior distribution over symbolic Gaussian
process model structures and numeric parameters that can accurately
model a range of real-world datasets.
We introduced involutive MCMC rejuvenation moves
(\crefrange{sec:smc-imcmc-subtree-replace}{sec:smc-imcmc-detach-attach})
that adapt the structure given new observations
and a new pseudo-marginal parameter proposal
(\cref{sec:smc-imcmc-pseudo-marginal}) to improve acceptance
rates of transdimensional moves.
These algorithmic contributions may also be of independent interest
for Bayesian structure learning in many other symbolic probabilistic
model families.

Empirical measurements in \cref{sec:evaluation-runtime}
show that our SMC sampler enables 10x--100x
improvements in runtime versus accuracy profiles over previous
greedy search and MCMC structure learning methods that target the same model class.
Our algorithms enabled us to perform in \cref{sec:evaluation-m3} a
large evaluation containing $\sim$1,400 forecasting problems drawn
from the challenging M3 dataset, which show that our method can
produce commonsense forecasts whose accuracy outperforms prominent
statistical and neural baselines across multiple forecast horizons.
These results present the first illustration of the practicality of
Gaussian process structure learning on a large benchmark set that is
challenging for both statistical and machine learning
methods~\citep{ahmed2010,makridakis2018,makridakis2023}.

As our structure learning algorithm operates over dense Gaussian
processes, it may be possible to work around the fundamental $O(n^3)$
scaling bottleneck by using sparse
approximations~\citep{quinonero2005,rossi2021}, state-space model
representations~\citep{hartikainen2010,Grigorievskiy2016}, or
combinations thereof~\citep{tebbutt2021,hamelijnck2021}.
These modeling variants would enable scalability beyond 10,000
observations at the expense of approximation accuracy, however,
introducing new trade-offs in the modeling and inference algorithm
design space.

\bibliography{paper}
\bibliographystyle{icml2023}


\newpage
\appendix
\onecolumn

\section{Proofs}
\label{appx:proofs}

\begin{proof}[\proofname\ of \Cref{eq:eta-gibbs}]

Let $(k, \theta, \eta, \mathbf{y})$ be the current state.
The conditional distribution of $[f(t_1), \dots, f(t_n)]$ is given by:
\begin{align}
[f(t_1), \dots, f(t_n)] \mid (k, \theta, \eta, \mathbf{y})
  &\sim \mathrm{MultivariateNormal}(\bm{\mu}, \bm{\Sigma}),\\
\bm{\mu} &\defas k_\theta(\mathbf{t})\left[k_\theta(\mathbf{t}) + \eta I\right]^{-1}\mathbf{y} \\
\bm{\Sigma} &\defas
  k_\theta(\mathbf{t}) -
  k_\theta(\mathbf{t})\left[k_\theta(\mathbf{t}) + \eta I\right]^{-1}
  k_\theta(\mathbf{t}).
\end{align}
Conditioning on $f(\mathbf{t}) = \bm{\mu}$
and applying conjugacy of the inverse-gamma prior~\labelcref{eq:GP-prior-3} to the normal
likelihood~\labelcref{eq:GP-prior-5} gives
\begin{align}
\eta \mid (k, \theta, f(\mathbf{t}) = \bm{\mu}, \mathbf{y})
  \sim \mathrm{InverseGamma}\left(1 + n/2, 1 + \sum_{i=1}^{n}(y_i - \mu_i)^2/2 \right).
\end{align}
\end{proof}

\begin{proof}[Proof of \cref{prop:subtree-replace}]
\label{proof:subtree-replace}
We will first show that $f_{\rm R}$ is an involution.
Let $(k, \theta, \eta)$ be the current state and
$(b, \tilde{k}, \tilde{\theta}, \eta') \sim q_{\rm R}(\cdot; k, \theta, \eta)$.
We then have
\begin{align}
f_{\rm R}([(k,\theta,\eta), (b, \tilde{k}, \tilde{\theta}, \eta')])
= [(k', \theta', \eta'), (b, \hat{k}, \hat{\theta}, \eta)],
&& \mbox{where } (k', \theta') \defas \otimes_b\left( \ominus_b(k, \theta), (\tilde{k}, \tilde{\theta}) \right)
\end{align}
The self-inversion property holds for the variables $(b, \eta, \eta')$,
because they each undergo a swap-position
operation under $f_{\rm R}$. It remains to show that the kernel structure
$k$ and parameters $\theta$ are self-inverted, that is,
\begin{align}
\otimes_b\left(
  \ominus_b \left(
    \otimes_b\left( \ominus_b(k, \theta), (\tilde{k}, \tilde{\theta}) \right)
    \right),
  (\hat{k}, \hat{\theta}) \right) = (k, \theta),
\label{eq:Columbid}
\end{align}
which follows because $(\hat{k}, \hat{\theta}) \defas (k_b, \theta_b)$
is the subtree of $(k, \theta)$ rooted at $b$ detached by the
innermost $\ominus_b$ operation.

To derive the acceptance ratio~\labelcref{eq:subtree-replace-accept},
the assumption that $q_{\rm R}$ samples the path $b$ uniformly and
$(\tilde{k},\tilde{\theta})$ from the conditional prior
given $({\ominus}_b(k, \theta))$ means that
\begin{align}
r_{\rm R}
&=
\frac
  {P_{\mathbf{t}}(k', \theta', \eta', \mathbf{y})}
  {P_{\mathbf{t}}(k, \theta, \eta, \mathbf{y})}
 \frac
  {q_{\rm R}(b, \hat{k}, \hat{\theta}, \eta; k', \theta', \eta', \mathbf{y})}
  {q_{\rm R}(b, \tilde{k}, \tilde{\theta}, \eta'; k, \theta, \eta, \mathbf{y})}
\cancelto{1}{
  \abs{
  \mathrm{Det} \
    J_{f_{\rm R}}(k,\theta,\eta,b,\tilde{k},\tilde{\theta},\eta')
  }}
\\
&=
\frac
  {P(\ominus_b(k', \theta')) P(\tilde{k}, \tilde{\theta} \mid \ominus_b(k', \theta'))}
  {P(\ominus_b(k, \theta)) P(\hat{k}, \hat{\theta} \mid \ominus_b(k, \theta))}
\frac
  {P(\eta')}
  {P(\eta)}
\frac
  {P_{\mathbf{t}}(\mathbf{y} \mid k', \theta', \eta')}
  {P_{\mathbf{t}}(\mathbf{y} \mid k', \theta', \eta')}
\frac
  {1/\abs{k'}}
  {1/\abs{k}}
\frac
  {P( \hat{k}, \hat{\theta} \mid \ominus_b (k', \theta'))}
  {P( \tilde{k}, \tilde{\theta} \mid \ominus_b (k, \theta))}
\frac
  {P_{\mathbf{t}}(\eta \mid \bm{\mu}, \mathbf{y})}
  {P_{\mathbf{t}}(\eta' \mid \bm{\mu}', \mathbf{y})}
\label{eq:seafowl}
\\
&=\frac{P(\eta')}{P(\eta)}
\frac
  {P_{\mathbf{t}}(\mathbf{y} \mid k',\theta',\eta')}
  {P_{\mathbf{t}}(\mathbf{y} \mid k,\theta,\eta)}
\frac{1/\abs{k'}}{1/\abs{k}}
\frac
  {P_{\mathbf{t}}(\eta \mid \bm{\mu}, \mathbf{y})}
  { P_{\mathbf{t}}(\eta' \mid \bm{\mu}', \mathbf{y})},
\label{eq:ammonolyze}
\end{align}
where \cref{eq:seafowl} uses the equality
$\ominus_b (k, \theta) \equiv \ominus_b (k', \theta')$.
\end{proof}

\begin{proof}[\proofname\ of \cref{prop:attach-detach}]
We first describe the implicit involution $f_{\rm DA}$
and proposal $q_{\rm DA}$ for \textsc{Attach-Detach}.
The proposal $q_{\rm DA}$ first samples
an auxiliary ``direction'' variable $d \in \set{0,1}$
\citep[Trick 3]{neklyudov2020} with probability $(\xi, 1 - \xi)$,
where $d=0$ corresponds to the $\textsc{Detach}$ direction and $d=1$ to $\textsc{Attach}$
direction.
Letting $\mathbf{z}$ be the current state and $\mathbf{v}$ be the
variables sampled by the proposals ($q_{\rm D}$ or $q_{\rm A}$),
let
$q_{\rm DA}(d=0, \mathbf{v}) \defas \xi q_{\rm D}(\mathbf{v})$,
$q_{\rm DA}(d=1, \mathbf{v}) \defas (1-\xi) q_{\rm A}(\mathbf{v})$.
Then
\begin{align}
f_{\rm DA}(d=0, \mathbf{z}, \mathbf{v}) &\defas (1, f_{\rm D}(\mathbf{z}, \mathbf{v})), \\
f_{\rm DA}(d=1, \mathbf{z}, \mathbf{v}) &\defas (0, f_{\rm A}(\mathbf{z}, \mathbf{v})).
\end{align}
To establish that $f_{\rm DA}$ is an involution, it suffices to show that
$f_{\rm D} = f_{\rm A}^{-1}$.
As in the previous proof, the variables $(a, b, \eta, \eta')$ each
undergo a swap-position operation under application of $f_{\rm A}, f_{\rm D}$
that invert one another.
It remains to be shown that
\begin{align}
\otimes_a(
  \ominus_a(
    \otimes_a (\ominus_a(k,\theta), (\tilde{k}, \tilde{\theta}))),
  (k_a, \theta_a)),
\end{align}
which follows from the properties
\begin{enumerate*}[label=(\roman*)]
\item $\ominus_a(\otimes_a(\ominus_a T, T')) = \ominus_a T$;
and
\item
$\otimes_a(\ominus_a T, T_a) = T$;
\end{enumerate*}
which hold for any tree $T$.

We now prove the expression for the acceptance ratio $r_{\rm D}$
corresponding to the \textsc{Detach} direction ($d = 0$).
Under the hypotheses in \cref{prop:attach-detach} and the fact that
$f_{\rm DA}$ is volume preserving, the acceptance ratio $r_{\rm A}$
is given by
\begin{align}
r_{\rm A}
&=
\frac
  {P_{\mathbf{t}}(k', \theta', \eta', \mathbf{y})}
  {P_{\mathbf{t}}(k, \theta, \eta, \mathbf{y})}
 \frac
  {q_{\rm DA}(d=1, a, b, \hat{k}, \hat{\theta}; k',\theta',\eta',\mathbf{y})}
  {q_{\rm DA}(d=0, a, b; k, \theta; k,\theta,\eta,\mathbf{y})}
\\
&=
\frac
  {P_{\mathbf{t}}(k', \theta', \eta', \mathbf{y})}
  {P_{\mathbf{t}}(k, \theta, \eta, \mathbf{y})}
 \frac
  {(1-\xi)}
  {\xi}
 \frac
  {q_{\rm A}(a,b,\hat{k},\hat{\theta}; k',\theta',\eta',\mathbf{y})}
  {q_{\rm A}(a,b; k,\theta,\eta,\mathbf{y})}
\\
&=
\begin{aligned}[t]
\frac
  {P(\ominus_a(k', \theta')) P(\tilde{k}, \tilde{\theta} \mid \ominus_a(k', \theta'))}
  {P(\ominus_a(k, \theta)) P(k_a, \theta_a \mid \ominus_a(k, \theta))}
\frac
  {(1-\xi)}
  {\xi}
\frac
  {P(\eta')}
  {P(\eta')}
&\frac
  {P_{\mathbf{t}}(\mathbf{y} \mid k', \theta', \eta')}
  {P_{\mathbf{t}}(\mathbf{y} \mid k', \theta', \eta')}
\frac
  {1/\abs{k'}}
  {1/\abs{k}}
\frac
  {q_{\rm A}(b \mid a; k', \theta')}
  {q_{\rm D}(b \mid a; k',\theta')} \\
&\frac
  {P(\hat{k}, \hat{\theta} \mid \ominus_a(k', \theta'), b, (\tilde{k},\tilde{\theta}))}
  {1}
\frac
  {P_{\mathbf{t}}(\eta \mid \bm{\mu}, \mathbf{y})}
  {P_{\mathbf{t}}(\eta' \mid \bm{\mu}', \mathbf{y})}
\end{aligned}
\\
\label{eq:seafowler}
&=\frac{P(\eta')}{P(\eta)}
\frac
  {P_{\mathbf{t}}(\mathbf{y} \mid k',\theta',\eta')}
  {P_{\mathbf{t}}(\mathbf{y} \mid k,\theta,\eta)}
\frac{1/\abs{k'}}{1/\abs{k}}
\frac
  {P_{\mathbf{t}}(\eta \mid \bm{\mu}, \mathbf{y})}
  { P_{\mathbf{t}}(\eta' \mid \bm{\mu}', \mathbf{y})}
\frac
  {(1-\xi)}
  {\xi}
\frac
  {q_{\rm A}(b \mid a; k', \theta')}
  {q_{\rm D}(b \mid a; k, \theta)},
\end{align}
where \cref{eq:seafowler} uses the following properties:
\begin{align}
\ominus_a (k, \theta) &= \ominus_a (k', \theta'), && \mbox{(by construction)} \\
P(\tilde{k}, \tilde{\theta} \mid \ominus_a (k', \theta')) &= P(\tilde{k}, \tilde{\theta}), && \mbox{(context-free prior)} \\
P(k_a, \theta_a \mid \ominus_a (k, \theta)) &= P(k_a, \theta_a), && \mbox{(context-free prior)} \\
P(k_a, \theta_a) &= P(\tilde{k},\tilde{\theta})P(k_a,\theta_a \mid ({k_a}_b,{\theta_a}_b) = (\tilde{k}, \tilde{\theta})) && \mbox{(chain rule)} \\
  ~              &= P(\tilde{k},\tilde{\theta})P(\hat{k}, \hat{\theta} \mid b, (\tilde{k}, \tilde{\theta})) && \mbox{(notation)}.
\end{align}
An identical argument establishes the expression for the acceptance ratio $r_{\rm D}$
of the \textsc{Attach} direction ($d = 1$).
\end{proof}

\begin{proof}[\proofname\ of~\cref{prop:subtree-replace-pseudo}]
Recalling the definition
of the involution $f^U_{\rm R}$~\labelcrefrange{eq:subtree-replace-inv1}{eq:subtree-replace-inv2}
for the \textsc{Subtree-Replace} proposal
on the extended state-space, the acceptance ratio is
\begin{align}
\tilde{r}^U_{\mathrm{R}}
&=\frac
  {P_{\mathbf{t}}(k', \theta', \eta', \mathbf{y})}
  {P_{\mathbf{t}}(k, \theta, \eta, \mathbf{y})}
\frac
  {q_{\rm R}(
    b, \hat{k}, (\theta^{1:\hat{u}-1}_{S,D},\theta_{S,D},\theta^{\hat{u}+1:U}_{S,D}),
    \hat{u}, (\theta'^{1:\tilde{u}-1}_{S,F},\theta^{\tilde{u}+1:U}_{S,F}), \tilde{u}, \eta)}
  {q_{\rm R}(b, \tilde{k}, \theta'^{1:U}_{S,F}, \tilde{u}, \theta^{1:U-1}_{S,D}, \hat{u}, \eta')}
\\
&=
\frac
  {P_{\mathbf{t}}(k', \theta', \eta', \mathbf{y})}
  {P_{\mathbf{t}}(k, \theta, \eta, \mathbf{y})}
\frac
  { 1/\abs{k} }
  { 1/\abs{k'} }
\frac
  {P(\hat{k} \mid \ominus_b k')}
  {P(\tilde{k} \mid \ominus_b k)}
\frac
  {q(\theta_{S,D})\prod_{u=1}^{U-1} q(\theta^u_{S,D})}
  {\prod_{u=1}^U q(\theta'^u_{S,F})}
\frac
  {\displaystyle \frac{\omega(\theta_{S,D})}{\sum_{u=1}^U \omega(\theta^u_{S,D})}}
  {\displaystyle \frac{\omega(\theta'^{\tilde{u}}_{S,F})}{\sum_{u=1}^U \omega(\theta'^u_{S,F})}}
\frac
  {\prod_{\substack{u=1\\ u\ne \tilde{u}}}^{U} q(\theta'^u_{S,F})}
  {\prod_{u=1}^{U-1} q(\theta^u_{S,D})}
\frac
  {1/U}
  {1/U}
\frac
  {P_{\mathbf{t}}(\eta \mid \bm{\mu}, \mathbf{y})}
  {P_{\mathbf{t}}(\eta' \mid \bm{\mu}', \mathbf{y})}
 \\
&=
\frac
  {P(\theta'^{\tilde{u}}_{S,F})}
  {P(\theta_{S,D})}
\frac
  {P(\eta')}
  {P(\eta)}
\frac
  {P_{\mathbf{t}}(\mathbf{y} \mid k', \theta', \eta')}
  {P_{\mathbf{t}}(\mathbf{y} \mid k, \theta, \eta)}
\frac
  { 1/\abs{k} }
  { 1/\abs{k'} }
\frac
  {q(\theta_{S,D})}
  {q(\theta'^{\tilde{u}}_{S,F})}
\frac
  {\omega(\theta_{S,D})}
  {\omega(\theta'^{\tilde{u}}_{S,F})}
\frac
  {\sum_{u=1}^U \omega(\theta'^u_{S,F})}
  {\sum_{u=1}^U \omega(\theta^u_{S,D})}
\frac
  {P_{\mathbf{t}}(\eta \mid \bm{\mu}, \mathbf{y})}
  {P_{\mathbf{t}}(\eta' \mid \bm{\mu}', \mathbf{y})}.
\label{eq:portability}
\end{align}
It can be readily verified that if $U = 1$ and $\delta = 0$ in the log-normal
proposals~\labelcref{eq:sir-11,eq:sir-21}, then \cref{eq:portability}
recovers \cref{eq:ammonolyze}, as these conditions
give $\theta_S \equiv \theta'^{\tilde{u}}_{S}$
which in turn implies that
$P(\theta'^{\tilde{u}}_{S,F}) q(\theta_{S,D}) = P(\theta_{S,D}) q(\theta'^{\tilde{u}}_{S,F})$.
To show that $f^U_{\rm R}$ is an involution, we apply it twice to the
current state to obtain
\begin{align}
&f^U_{\mathrm{R}}\left(
  f^U_{\mathrm{R}}([
    (k, \theta, \eta),
    (b, \tilde{k}, \theta'^{1:U}_{S,F}, \tilde{u}, \theta^{1:U-1}_{S,D}, \hat{u}, \eta')
    ])\right)
    \\
&=f^U_{\mathrm{R}}\left(
  \left [
    \left(
      \otimes_b\left( \ominus_b(k, (\theta'^{\tilde{u}}_S, \theta_D)), (\tilde{k}, {\theta'^{\tilde{u}}_F}) \right),
      \eta'
      \right),
    (b, \hat{k},
      (\theta^{1:\hat{u}-1}_{S,D},\theta_{S,D},\theta^{\hat{u}:U-1}_{S,D}),
      \hat{u},
      (\theta'^{1:\tilde{u}-1}_{S,F},\theta'^{\tilde{u}+1:U}_{S,F}),
      \tilde{u},
      \eta)
    \right]
  \right) \\
&=
  \left [
    (
      \underbrace{k, \left(\theta_S, \theta_D\right)}_{\mathrm{\cref{eq:Columbid}}},
      \eta
      ),
    (b, \tilde{k},
      (\theta^{1:\tilde{u}-1}_{S,F},\theta'^{\tilde{u}}_{S,F},\theta'^{\tilde{u}:U-1}_{S,F}),
      \tilde{u},
      (\theta^{1:\hat{u}-1}_{S,D},\theta^{\hat{u}:U-1}_{S,D}),
      \hat{u},
      \eta')
    \right].
\end{align}

The transition kernel with acceptance probability
$\alpha^U_{\rm R} = \min(1, \tilde{r}^U_{\mathrm{R}})$
thus leaves $P_{\mathbf{t}}(k,\varphi\,{\mid}\,\mathbf{y})$ invariant.
Irreducibility and aperiodicity follow from
\begin{enumerate*}[label=(\roman*)]
\item the hypothesis that
$q^U_{\rm R}$ selects a zero-length path $b$ (i.e., the root node) with positive probability; and
\item the full support of the proposal $q(\theta'; \theta)$ over $\Theta_{d(k')}$ and of the proposal
 $P(\eta' \mid \bm{\mu}', \mathbf{y})$ over $\mathbb{R}_{>0}$.
\end{enumerate*}
These conditions guarantee that any positive probability set in the
transdimensional space~\labelcref{eq:target-support} can be reached
from $(k, \theta, \eta)$ in one step.
\end{proof}

\section{Profiles of Effective Sample Size vs.~SMC Step}
\label{appx:ess}

\begin{figure*}[!t]
\centering
\includegraphics[width=\linewidth]{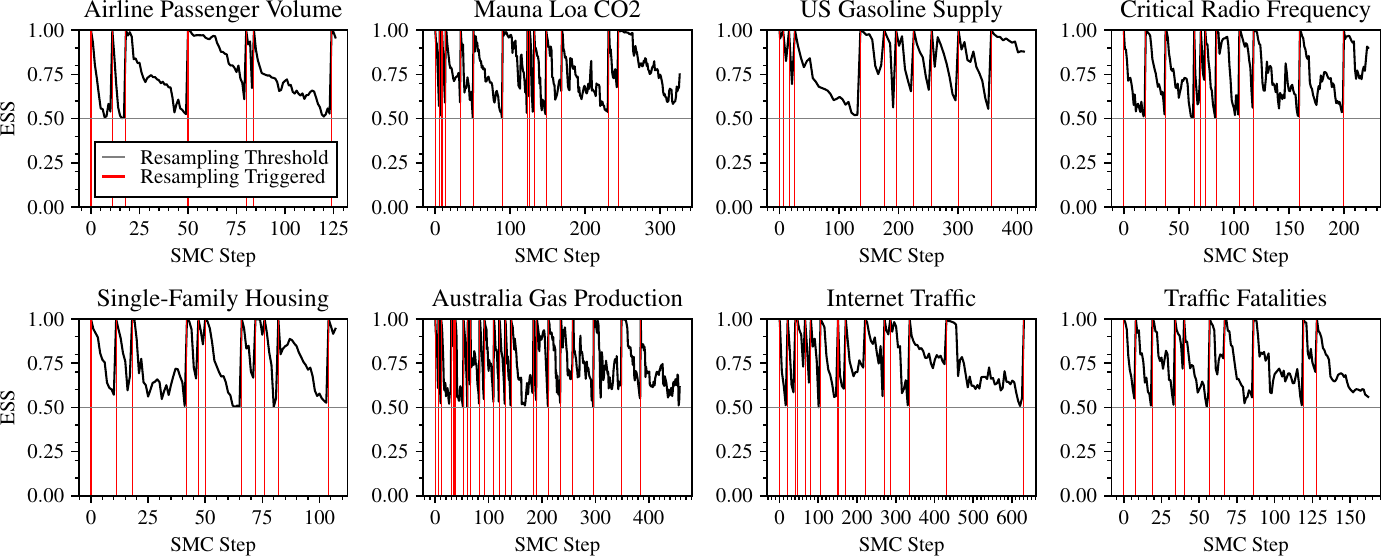}
\captionsetup{belowskip=-10pt}
\caption{SMC step (x-axes) vs.~normalized effective sample
size (y-axes, ESS) for the eight datasets used in~\cref{fig:runtime}.}
\label{fig:ess}
\end{figure*}

\Cref{fig:ess} shows the normalized effective sample size (ESS) versus SMC step $j$,
defined from the weights according to
$\mathrm{ESS}_j \defas (\sum_{i=1}^M(w^i_j))^2 / (M\sum_{i=1}^M (w^i_j)^2)$,
for eight datasets analyzed in \cref{fig:runtime}
(\cref{sec:evaluation-runtime} of the main text).
Horizontal lines at ESS = 0.5 show the resampling threshold and
vertical red lines show steps at which resampling occurred.
There are typically several rejuvenation steps between resampling steps,
suggesting that particle degeneracy is not a significant issue.

\section{M3 Evaluation Details}
\label{appx:evaluation}

\Cref{table:baselines} enumerates the baselines used in the evaluation
from \cref{sec:evaluation-m3}.\footnote{
We also considered the hand-designed Gaussian
process model from~\citet{corani2021}, which uses a fixed structure with
five kernels \textsc{(Periodic + Linear + RBF + SM$_1$ + SM$_2$)}
tailored to the M3 data, where \textsc{SM} is a spectral mixture kernel.
However, we encountered two challenges:
\begin{enumerate*}[label=(\roman*)]
\item the model hyperpriors are pre-tuned on 350/1428
monthly time series, so the forecasts apply only to 1078/1428 datasets; and
\item the prediction intervals spanned many orders of magnitude,
leading to MSIS errors of ${\approx}\,1.8{\times}{10}^4$.
\end{enumerate*}}
For each method, the default inference settings were used where applicable.
Baselines from the statsforecast and neuralforecast packages
\citep{garza2022} require the seasonal period $m$ as input, which was
set as $m=12$ for monthly data.
The LSTM and Temporal Fusion Transformer were trained with the
multi-quantile loss at level 95 to produce the required 95\% forecast
intervals.
For Singular Spectrum Analysis (SSA), multi-horizon forecasts were
obtained using the recurrent SSA forecasting algorithm with bootstrap
estimation for the forecast intervals.
For \AutoGP{}, \Cref{alg:smc} was run using a linear annealing
schedule with 5\% of the data introduced at each step; $M=48$
particles; adaptive resampling with $\mathrm{ESS} = M/2 = 24$; and
$N_{\rm rejuv} = 100$ MCMC rejuvenation steps.
The training time points $\mathbf{t}$ and (demeaned) values
$\mathbf{y}$ are linearly transformed to $[0,1]$.
We use a kernel language $\mathcal{L}$ with
three base kernels $B$ (\cref{eq:pcfg-1}) and a changepoint
operator (\textsc{CP}, \cref{eq:pcfg-2}) with the following
parameterization:
\begin{align}
\textsc{Linear}_{\theta_1, \theta_2, \theta_3}
  &\defas \lambda t, t'. \theta_1 + \theta_2(t - \theta_3)(t' - \theta_3)
\\
\textsc{GammaExponential}_{\theta_1, \theta_2, \theta_3}
  &\defas \lambda t, t'. \theta_1 \exp\left( - (\abs{t-t'}/\theta_2)^{\theta_3} \right) && \theta_3 \in (0, 2]
\\
\textsc{Periodic}_{\theta_1, \theta_2, \theta_3}
  &\defas \lambda t, t'. \theta_1 \exp\left((-2/\theta_2^2)  (\sin(\pi \abs{t-t'} / \theta_3 ))^2\right)
\\
k_1\ \textsc{CP}_{\theta_1, \theta_2}\ k_2
  &\defas \lambda t, t'.
    \begin{aligned}[t]
    &[\sigma_1 \cdot k_1(t, t') \cdot \sigma_2]
    +
    [(1 - \sigma_1) \cdot k_2(t, t') \cdot (1-\sigma_2)]\\
    &\mbox{where }
      \begin{aligned}[t]
        \sigma_1 &\defas (1 + \tanh((t - \theta_1) / \theta_2))/2 \\
        \sigma_2 &\defas (1 + \tanh((t' - \theta_1) / \theta_2))/2
      \end{aligned}
    \end{aligned}
\end{align}

\begin{table}
\caption{Baseline methods used in the evaluation from \cref{sec:evaluation-m3} on the M3 monthly time series.}
\label{table:baselines}
\footnotesize
\begin{tabular*}{\linewidth}{@{\extracolsep{\fill}}lll}
\toprule
Method & Reference & Implementation \\ \midrule
AutoARIMA & \citet{hyndman2008ets} & \url{https://nixtla.github.io/statsforecast/} \\
AutoETS & \citet{hyndman2008ets} & \url{https://nixtla.github.io/statsforecast/} \\
AutoTheta & \citet{fiorucci2016} & \url{https://nixtla.github.io/statsforecast/}\\
Custom M3 Gaussian Process & \citet{corani2021} & \url{https://github.com/IDSIA/gpforecasting/} \\
Facebook Prophet & \citet{taylor2018} & \url{https://facebook.github.io/prophet/} \\
LSTM & \citet{lindemann2021} & \url{https://nixtla.github.io/neuralforecast/}\\
Random Walk with Drift & \citet{hyndman2021} & \url{https://nixtla.github.io/statsforecast/} \\
Seasonal Naive & \citet{hyndman2021} & \url{https://nixtla.github.io/statsforecast/} \\
Singular Spectrum Analysis & \citet{golyandina2018} & \url{https://cran.r-project.org/web/packages/Rssa/} \\
Temporal Fusion Transformer & \citet{lim2021} & \url{https://nixtla.github.io/neuralforecast/} \\
\bottomrule
\end{tabular*}
\end{table}

\subsection{Performance Metrics for Evaluating Forecasts}
\label{appx:evaluation-metrics}

The definitions of Symmetric Mean Absolute Percentage Error (SMAPE),
Mean Absolute Scaled Error (MASE), and Mean Scaled Interval Score
(MSIS) appearing in \cref{fig:runtime,fig:m3-eval} are taken from
\citet{makridakis2018}:
\begin{align}
\label{eq:smape}
\mathrm{SMAPE}_h(x, \hat{x}) &\defas
100 \times \frac{2}{h}\sum_{t=n+1}^{n+h} \frac{\abs{x_t - \hat{x}_t}}{\abs{x_t} + \abs{\hat{x}_t}}
\\
\label{eq:mase}
\mathrm{MASE}_{h,m}(x, \hat{x}) &\defas
\frac{1}{h}
\frac{\sum_{t=n+1}^{n+h} \abs{x_t - \hat{x}_t}}
{\displaystyle\frac{1}{n-m} \sum_{t=m+1}^{n}\abs{x_t - x_{t-m}}}
\\
\label{eq:msis}
\mathrm{MSIS}_{h,m}(x, \hat{u}, \hat\ell) &\defas
\frac{1}{h}
\frac{
    \sum_{t=n+1}^{n+h} (\hat{u}_t - \hat{\ell}_t)
    +  \frac{2}{a}(\hat{\ell}_t-x_t)\mathbf{1}[x_t < \hat{\ell}_t]
    +  \frac{2}{a}(x_t < \hat{u}_t)\mathbf{1}[\hat{u}_t < x_t]}
{\displaystyle\frac{1}{n-m} \sum_{t=m+1}^{n}\abs{x_t - x_{t-m}}},
\end{align}
where $n$ is the number of observed data points; $h > 0$ is the
forecasting horizon; $m > 0$ is the seasonal period ($m=12$ for
monthly data); $x = (x_1, \dots, x_n, x_{n+1}, \dots, x_{n+h})$
denotes the $n$ observed data points and $h$ test data points;
$\hat{x} = (\hat{x}_{n+1}, \dots, \hat{x}_{n+h})$ denotes the $h$ point forecasts;
and $\hat{u} = (\hat{u}_{n+1}, \dots, \hat{u}_{n+h})$ and $\hat{\ell} =
(\hat{\ell}_{n+1}, \dots, \hat{\ell}_{n+h})$ denote the $h$ upper and
lower bounds of the $(1-a)\%$ prediction interval ($a=0.05$ for 95\%
intervals).

\section{Modeling Discrete-Time Data with Gaussian Processes}

Recall from \cref{eq:GP-prior-4,eq:GP-prior-5} that the Gaussian
process time series model specifies a distribution over data
$y_i \sim \Dist{Normal}(f(t_i), \eta)$ where
$t_i \in \mathbb{R}$ is a real time index.
For discrete-time datasets such as M3, the observed
time index $t_i$ is instead a time stamp of the form YYYY-MM-DD.
We convert time stamps to numbers by computing their age in seconds since
the UNIX epoch 1970-01-01 00:00:00.
This strategy is preferable to converting time stamps to integer indexes
in applications where even slight differences in the duration between
two time stamps influences the magnitude of the
observations~\citep{loossens2021}.
In a monthly time series, for example, the month
2020-01-01---2020-02-01 contains 31 days while the month
2020-02-01---2020-03-01 contains only 29 days.
The UNIX epoch encoding of these time stamps
(1577836800, 1580515200, 1583020800) captures the unequal spacing
whereas the integer encoding (1, 2, 3) does not.
Another benefit of epoch encoding with continuous-time models is that it
provides a coherent way of learning seasonal periods
with non-integer periodicity (such as weekly data with an annual
pattern which has a seasonal period of 365.25/7 $\approx$ 52.179 steps
on average).
Traditional ARIMA or ETS models do not handle non-integer
periodicity~\citep[Section 12.1]{hyndman2021} and require users to
manually include Fourier features with the correct periodicity.

Caution must be taken when computing predictions with a Gaussian
process trained on epoch encodings, however, because the time series
may not have meaningful semantics at arbitrary time points.
Consider, for example, a monthly time series that records the total number of
home sales as measured on the final day (2020-01-31, 2020-02-28, \dots).
The output $f(t)$ of a Gaussian process $f$ at a real number $t$ that
encodes an arbitrary time stamp (e.g., 2020-01-18, 8pm) may not
correspond to a semantically well-defined measurement of the
underlying time series.
Time series that evolve in continuous time can generally be modeled
without these subtleties.

\end{document}